\RequirePackage{fix-cm}
\documentclass[twocolumn]{svjour3}          
\smartqed  
\usepackage{graphicx}
\usepackage{amsmath,amssymb}
\usepackage{mathptmx}
\usepackage{multirow}
\usepackage{marvosym}
\usepackage{booktabs}
\usepackage{subcaption}
\usepackage{bm}
\usepackage{rotating}
\usepackage{enumitem}
\usepackage{color}
\usepackage[pagebackref=true, colorlinks,bookmarks=false]{hyperref}

\begin{document}

\title{RigNet++: Semantic Assisted Repetitive Image Guided Network for Depth Completion}


\author{Zhiqiang Yan\textsuperscript{1} \and Xiang Li\textsuperscript{2} \and Le Hui\textsuperscript{3} \and Zhenyu Zhang\textsuperscript{4} \and Jun Li\textsuperscript{1}
\and Jian Yang\textsuperscript{1}
\\
{\small \textsuperscript{1}{\{yanzq, junli, csjyang\}@njust.edu.cn}} ~~{\small \textsuperscript{2}xiang.li.implus@nankai.edu.cn} ~~{\small \textsuperscript{3}huile@nwpu.edu.cn} ~~{\small \textsuperscript{4}zhenyuzhang@nju.edu.cn}}


\institute{
\textsuperscript{1} PCA Lab, 
Nanjing University of Science and Technology, China.\\
\textsuperscript{2} Nankai University, China.\\
\textsuperscript{3} Northwestern Polytechnical University, China.\\
\textsuperscript{4} Nanjing University, China.}
\authorrunning{Zhiqiang Yan, Xiang Li, Le Hui, Zhenyu Zhang, Jun Li and Jian Yang}

\date{Received: date / Accepted: date}

\maketitle

\begin{abstract}
Depth completion aims to recover dense depth maps from sparse ones, where color images are often used to facilitate this task. 
Recent depth methods primarily focus on image guided learning frameworks. However, \emph{blurry guidance in the image} and \emph{unclear structure in the depth} still impede their performance. To tackle these challenges, we explore a repetitive design in our image guided network to gradually and sufficiently recover depth values. Specifically, the repetition is embodied in both the image guidance branch and depth generation branch. In the former branch, we design a dense repetitive hourglass network (DRHN) to extract discriminative image features of complex environments, which can provide powerful contextual instruction for depth prediction. In the latter branch, we present a repetitive guidance (RG) module based on dynamic convolution, in which an efficient convolution factorization is proposed to reduce the complexity while modeling high-frequency structures progressively. Furthermore, in the semantic guidance branch, we utilize the well-known large vision model, \emph{i.e.}, segment anything (SAM), to supply RG with semantic prior. In addition, we propose a region-aware spatial propagation network (RASPN) for further depth refinement based on the semantic prior constraint. Finally, we collect a new dataset termed TOFDC for the depth completion task, which is acquired by the time-of-flight (TOF) sensor and the color camera on smartphones. Extensive experiments demonstrate that our method achieves state-of-the-art performance on KITTI, NYUv2, Matterport3D, 3D60, VKITTI, and our TOFDC. 
\end{abstract}

\section{Introduction}
Depth completion, the technique of converting sparse depth measurements to dense ones, has a variety of applications in computer vision field, such as autonomous driving \cite{hane20173d,cui2019real,wang2021regularizing,yan2023desnet,wang2023lrru,yu2023aggregating}, augmented reality \cite{dey2012tablet,song2020channel,yan2022learning}, virtual reality \cite{armbruster2008depth,yan2023distortion,chen2023agg}, and 3D scene reconstruction \cite{park2020nonlocal,shen2022panoformer,yan2022multi,zhou2023bev}. The success of these applications heavily depends on reliable depth predictions. Recently, multi-modal information from various sensors is involved to help generate dependable depth results, such as color images \cite{chen2019learning,yan2022rignet}, surface normals \cite{zhang2019pattern,Qiu_2019_CVPR}, confidence maps \cite{2020Confidence,vangansbeke2019}, and even binaural echoes \cite{gao2020visualechoes,parida2021beyond}. Particularly, the latest image guided methods \cite{liu2023mff,yan2022rignet,lin2022dynamic,zhang2023cf} mainly focus on using color images to guide the recovery of dense depth maps, achieving outstanding performance. However, due to the challenging environments and limited depth measurements, it is usually difficult for existing approaches to produce clear image guidance and structure-detailed depth features. To deal with these issues, in this paper we develop a repetitive design in both the image guidance branch and depth generation branch, while in the semantic guidance branch we employ the large vision model SAM \cite{kirillov2023segment} to provide semantic prior. 

\begin{figure*}[t]
\centering
\includegraphics[width=1.8\columnwidth]{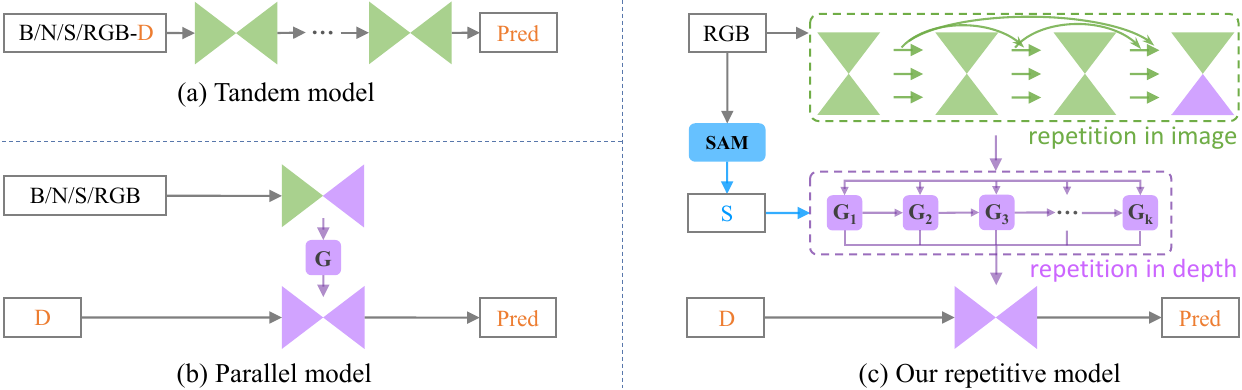}
\caption{To obtain dense depth \textbf{Pred}iction, most existing image guided methods employ (a) tandem models~\cite{ma2018self,Cheng2020CSPN,park2020nonlocal,lin2022dynamic,zhang2023cf} or (b) parallel models ~\cite{zhao2021adaptive,tang2020learning,liu2021fcfr,hu2020PENet,liu2023mff} with various inputs, \emph{e.g.}, \textbf{B}oundary, \textbf{N}ormal, \textbf{S}emantic, and RGB-D. By contrast, we propose (c) dense repetitive mechanism, which is assisted by the semantic prior of large vision model, \emph{i.e.}, segment anything (SAM) \cite{kirillov2023segment}, to gradually produce refined image and depth \textbf{G}uidance with rich semantic information.
}\label{model_summary}
\end{figure*}

In the image guidance branch: Existing image guided methods are not sufficient to produce very precise details to provide perspicuous image guidance, which limits the content-complete depth recovery. For example, the tandem models in Fig.~\ref{model_summary}(a) tend to utilize only the last layer features of one hourglass unit to guide the depth prediction. The parallel models in Fig.~\ref{model_summary}(b) either have scarce interaction among multiple hourglass units, or rely on image guidance encoded by only a single hourglass unit. Different from these models, as shown in Fig.~\ref{model_summary}(c), we present a vertically dense repetitive hourglass network (DRHN) that can make good use of color image features in multi-scale layers, contributing to much clearer and richer contexts. However, the repetition in DRHN is quite complex. To reduce the complexity, we utilize some well-known lightweight backbones \cite{tan2019efficientnet,tan2021efficientnetv2}, and introduce a dense-connection strategy from a new perspective that differs from DenseNet \cite{huang2017densely}. 

In the depth generation branch: It is known that gradients near boundaries usually have large mutations, which increase the difficulty of recovering structure-detailed depth for convolution \cite{Uhrig2017THREEDV}. As evidenced in plenty of methods \cite{huang2019indoor,2020Confidence,park2020nonlocal},  the depth values are usually hard to be predicted especially around the region with unclear boundaries. To moderate this issue, we propose a repetitive guidance (RG) module based on dynamic convolution \cite{tang2020learning}. It first extracts the high-frequency components by channel-wise and cross-channel filtering, and then repeatedly stacks the guidance unit to progressively produce refined depth. We also design an adaptive fusion mechanism to effectively obtain better depth representations by aggregating depth features of each repetitive unit. However, an obvious drawback of the dynamic convolution is the large GPU memory consumption, especially under the case of our repetitive structure. Hence, we further introduce an efficient module to largely reduce the memory cost but maintain the accuracy. 

In the semantic guidance branch: With the rapid growth of data and computing power, large-scale vision models such as SAM \cite{kirillov2023segment} have become prevalent. Recently, more and more deep learning methods \cite{mazurowski2023segment,chen2023sam,lin2023sam,yu20233d,li2023sam} have started to leverage the prior knowledge of these large models for task-specific facilitation. Inspired by them, we use SAM to produce high-level semantic information, which can also provide coarse structure guidance through its mask edges. As a result, the image features in the image guidance branch, along with the semantic features in the semantic guidance branch, are fed into RG to guide the depth generation. 

For further structure refinement, we introduce a region-aware spatial propagation network (RASPN). Unlike previous SPN methods \cite{cspneccv,park2020nonlocal,liu2022graphcspn,lin2022dynamic,zhou2023bev} that acquire their affinitive neighbors with RGB-D cues, 
our RASPN embeds an additional semantic prior constraint. It explicitly restricts the SPN to search for its neighbors within a single-mask region. Therefore, the affinity preserves more accurate geometric structures, especially near the edges. 

In addition, since depth information plays a crucial role in accurate 3D reconstruction and human-computer interaction, time-of-flight (TOF) depth sensors are increasingly equipped on edge mobile devices. In this paper, we collect a new depth completion dataset termed TOFDC, with a smartphone that has both TOF lens and color camera. 

To summarize, our contributions are: 
(1) A novel architecture for depth completion is designed. 
(2) We present DRHN to extract clear image features, and RG, which consists of efficient guidance and adaptive fusion for precise depth recovery. 
(3) For the first time, we introduce the semantic prior of SAM to facilitate the depth completion task. 
(4) We propose RASPN, which incorporates the semantic prior constraint for further depth structure refinement. 
(5) We build TOFDC, a new smartphone-based depth completion dataset. Moreover, our method consistently outperforms the state-of-the-art approaches on six RGB-D datasets. 

This paper expands upon the initial version known as RigNet \cite{yan2022rignet}. The image guidance branch of RigNet, characterized by its complex repetition of multiple hourglass units, has limited feature representation due to the minimal interaction between these units. Additionally, the RG module in the depth generation branch lacks explicit high-level semantic guidance, leading to subpar depth structure recovery. To address these challenges, this paper introduces several improvements. Firstly, the initial hourglass units are replaced with efficient backbones, supplemented by a dense-connection strategy. Secondly, the semantic prior of SAM and a semantic constrained spatial propagation network are incorporated for enhanced depth structure generation. 
Compared to RigNet \cite{yan2022rignet}, the improvements of this paper are fourfold: 
(1) It reduces the complexity of the repetition in the image guidance branch and enhances its feature representation via the dense-connection strategy. 
(2) It introduces the semantic prior of SAM to benefit the repetition in the depth generation branch. 
(3) It proposes a region-aware spatial propagation network (RASPN) to refine the depth structure through explicit semantic prior constraints. 
(4) It constructs a new dataset based on the TOF system of a smartphone. 
(5) Additional experiments are performed on two panoramic RGB-D datasets and the TOFDC. 
These enhancements aim to address the limitations of RigNet and provide a more robust and efficient solution.

\section{Related Work}

\subsection{Depth Completion} 
\emph{Depth-only}: For the first time in 2017, the literature \cite{Uhrig2017THREEDV} proposes sparsity invariant CNNs to deal with sparse depth. Since then, plenty of depth completion methods \cite{Uhrig2017THREEDV,ku2018defense,2018Deep,2018Sparse,2020Confidence,ma2018self,vangansbeke2019} has been proposed, which only take sparse depth as input without using color image. For example, S2D \cite{ma2018self} employs an UNet model to predict dense depth. NConv \cite{2020Confidence} presents an algebraically-constrained normalized convolution layer to model sparse depth input. CU-Net \cite{wang2022cu} develops a local UNet and a global UNet to remove outliers and enhance sparse-to-dense depth prediction. Besides, Lu \emph{et al.} \cite{2020FromLu} employ color image as auxiliary supervision during training while only taking sparse depth as input when testing. However, single-modal based methods are limited without other reference information. As technology quickly develops, lots of multi-modal information is available, \emph{e.g.}, surface normal, semantic segmentation, and optic flow images, which can further facilitate the depth completion task. 

\emph{Image guided}: Existing image guided depth completion methods can be roughly divided into two patterns. One pattern is that various maps are together input into tandem hourglass networks \cite{cspneccv,chen2019learning,Cheng2020CSPN,park2020nonlocal,xu2020deformable,lin2022dynamic,yan2023desnet,zhang2023cf}. For example, S2D \cite{ma2018self} directly feeds the concatenation into a simple UNet \cite{ronneberger2015u}. CSPN \cite{cspneccv} studies the affinity matrix to refine coarse depth with spatial propagation network (SPN). CSPN++ \cite{Cheng2020CSPN} further improves its effectiveness and efficiency by learning adaptive convolutional kernel sizes and the number of iterations for propagation. As an extension, NLSPN \cite{park2020nonlocal} presents non-local SPN which focuses on relevant non-local neighbors during propagation. DSPN \cite{xu2020deformable} proposes a deformable SPN to adaptively generate different receptive field and affinity matrix at each pixel for effective propagation. DySPN \cite{lin2022dynamic} then develops a dynamic SPN via assigning different attention levels to neighboring pixels of different distances. The expanded version \cite{lin2023dyspn_tcsvt} of DySPN further introduces a dynamic path and diffusion suppression for better depth recovery. Different from these methods, FuseNet \cite{chen2019learning} designs an effective block to extract joint 2D and 3D features. BDBF \cite{Qu_2021_ICCV} predicts depth bases and computes basis weights for differentiable least squares fitting. On the basis of vision transformer \cite{dosovitskiy2020image}, GFormer \cite{rho2022guideformer} and CFormer \cite{zhang2023cf} concurrently leverage convolution and transformer to extract both local and long-range feature representations. 
Another pattern is using multiple independent branches to encode different sensor information and then fuse them at multi-scale stages \cite{vangansbeke2019,2020Denseyang,li2020multi,liu2021fcfr,hu2020PENet,yan2022rignet,yan2023learnable,chen2023agg}. For instance, DLiDAR \cite{Qiu_2019_CVPR} produces surface normals to improve model performance. PENet \cite{hu2020PENet} employs feature addition to guide depth learning at different stages. FCFRNet \cite{liu2021fcfr} proposes channel-shuffle technology to enhance RGB-D feature fusion. ACMNet \cite{zhao2021adaptive} chooses graph propagation to capture the observed spatial contexts. GuideNet \cite{tang2020learning} seeks to predict dynamic kernels from guided images to adaptively filter depth features. In addition, for the first time BEVDC \cite{zhou2023bev} introduces bird's-eye-view projection to leverage the geometric details of sparse measurements. Most recently, LRRU~\cite{wang2023lrru} presents a large-to-small dynamical kernel to capture long-to-short dependencies. \cite{yu2023aggregating}. PointDC \cite{yu2023aggregating} introduces a LiDAR point cloud branch to propagate 3D geometry. However, these methods still cannot provide very sufficient semantic guidance based on color images.

\subsection{Repetitive Learning Models} 
To extract more accurate and abundant feature representations, many approaches~\cite{ren2015faster,cai2018cascade,liu2020cbnet,wang2022cu,qiao2021detectors} propose to repeatedly stack similar components. For example, PANet \cite{liu2018path} adds an extra bottom-up path aggregation which is similar with its former top-down feature pyramid network (FPN). NAS-FPN \cite{ghiasi2019fpn} and BiFPN \cite{tan2020efficientdet} conduct repetitive blocks to sufficiently encode discriminative image semantics for object detection. FCFRNet \cite{liu2021fcfr} argues that the feature extraction in one-stage frameworks is insufficient, and thus proposes a two-stage model, which can be regarded as a special case of the repetitive design. On this basis, PENet \cite{hu2020PENet} further improves its performance by utilizing confidence maps and varietal CSPN++.
Different from these methods, in our image branch we first conduct the dense repetitive hourglass units to produce clearer guidance in multi-scale layers. Then in our depth branch we perform the efficient repetitive guidance module to generate structure-detailed depth prediction. 

 \begin{figure*}[t]
  \centering
  \includegraphics[width=1.87\columnwidth]{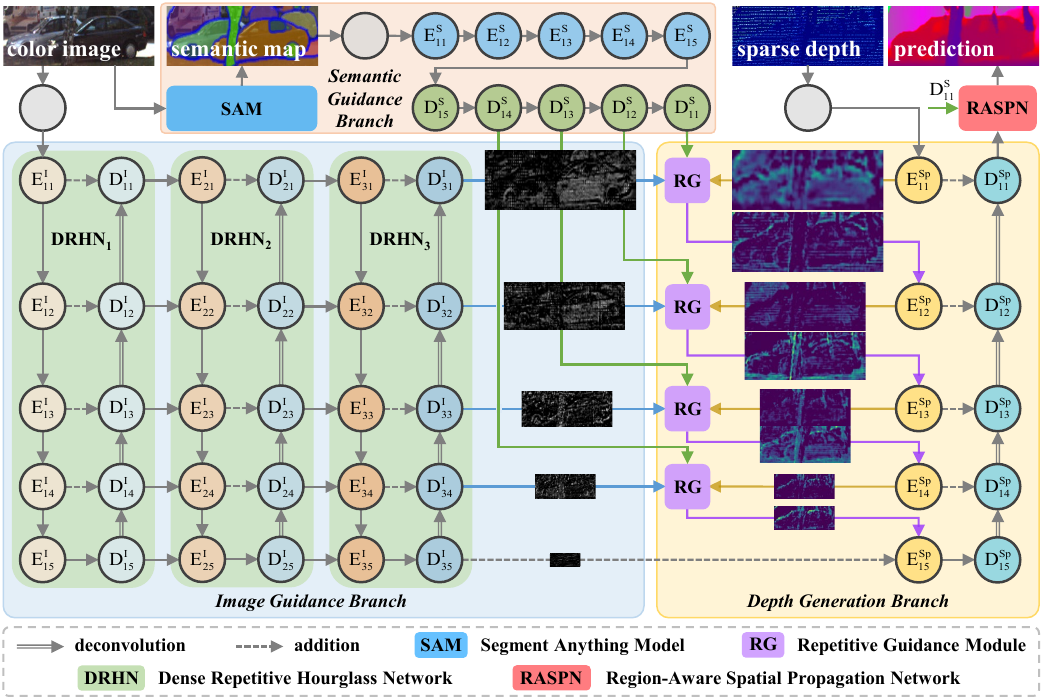}\\
  \caption{Overview of our semantic assisted repetitive image guided network. It mainly consists of image guidance branch, semantic guidance branch, and depth generation branch. Our RG (Fig.~\ref{Fig_RG_EG}) produces dense depth by fusing the features of the three branches, while the post-processing RASPN (Fig.~\ref{Fig_RASPN}) further refines the coarse depth using the semantic constraint. 
  }\label{Fig_pipeline}
\end{figure*}

\subsection{SAM Assisted Vision Tasks} 
Owing to the strong representation and generalization abilities, the large-scale vision model SAM \cite{kirillov2023segment} has found extensive applications across a diverse range of tasks, such as tracking \cite{yang2023track}, 6D pose estimation \cite{lin2023sam}, point cloud instance segmentation \cite{yu20233d}, image deblurring \cite{li2023sam}, shadow detection \cite{chen2023sam}, and medical image segmentation \cite{mazurowski2023segment,ma2023segment,huang2023segment,zhang2023customized}. These tasks leverage the segmentation results of SAM to promote their models' performance. Motivated by these successful applications, we introduce the semantic prior of SAM to provide explicit semantic guidance for the vanilla RigNet.

\section{RigNet++}
We present our network architecture in Sec.~\ref{sec_architecture} and our dense repetitive hourglass network in Sec.~\ref{DRHN}. Then, we explain how we use SAM in Sec.~\ref{semantic_guidance_branch} and how we design the repetitive guidance module in Sec.~\ref{RG}, which includes an efficient guidance algorithm and an adaptive fusion mechanism. Lastly, we introduce the region-aware spatial propagation network in Sec.~\ref{RASPN} and define the loss in Sec.~\ref{loss_func}. 

\subsection{Network Architecture}\label{sec_architecture}
Fig.~\ref{Fig_pipeline} shows the overview of RigNet++, including the image guidance branch, semantic guidance branch, depth generation branch, repetitive guidance (RG), and region-aware spatial propagation network (RASPN). 

\begin{figure*}[t]
  \centering
  \includegraphics[width=1.93\columnwidth]{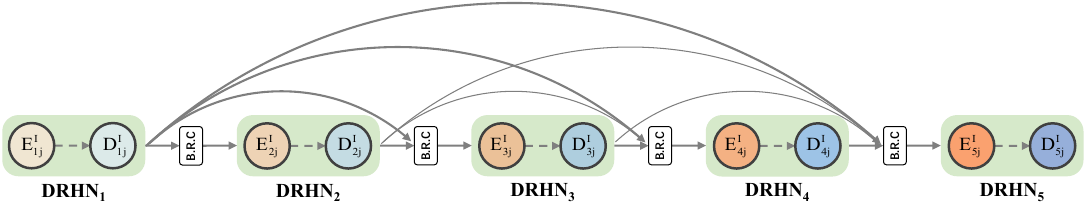}\\
  \vspace{-3pt}
  \caption{An example of our dense repetitive hourglass network (DRHN). B.R.C refers to BN, ReLU, and convolution.}\label{Fig_DRHN}
  \vspace{0pt}
\end{figure*}

The image guidance branch leverages a series of dense repetitive hourglass networks (DRHN) to produce hierarchical image features. DRHN$_1$  is a symmetric UNet, whose encoder is built upon the lightweight EfficientNet \cite{tan2019efficientnet} for high efficiency. Compared with DRHN$_1$, DRHN$_i$ ($i>1$) has the same or more lightweight architecture, which is used to extract clearer image guidance semantics \cite{zeiler2014visualizing}. In addition, we adopt the skip connection strategy \cite{ronneberger2015u,chen2019learning} to utilize low-level and high-level features simultaneously in DRHN$_i$. Between multiple DRHN subnetworks, a dense-connection strategy (see Fig.~\ref{Fig_DRHN}) is conducted. 

The semantic guidance branch contains a single symmetric UNet that is similar to DRHN$_1$. The color image is first fed into the large vision model SAM \cite{kirillov2023segment} to produce the high-level semantic segmentation image. Then the UNet maps it into feature space for further multi-modal fusion. 

\begin{figure*}[t]
  \centering
  \includegraphics[width=1.9\columnwidth]{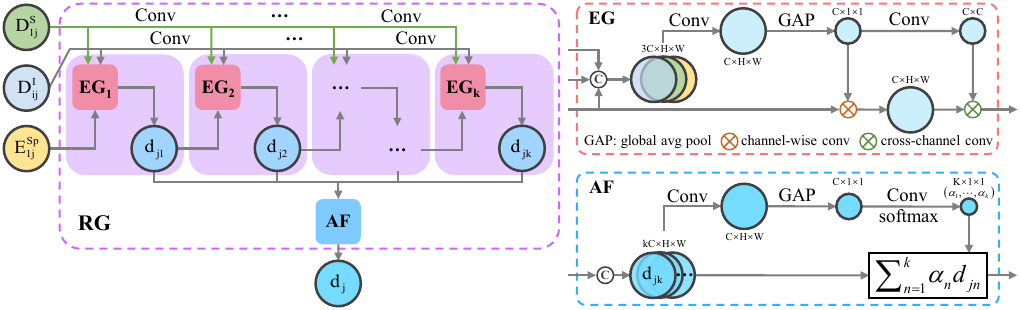}\\
  \caption{Our repetitive guidance (RG) module that consists of an efficient guidance algorithm (EG) and an adaptive fusion mechanism (AF), where $k$ refers to the repetitive number.}\label{Fig_RG_EG}
\end{figure*}

The depth generation branch has the same structure as DRHN$_1$. Based on the dynamic convolution \cite{tang2020learning}, we design the repetitive guidance (RG) module to gradually predict structure-detailed depth at multiple stages. Besides, an efficient guidance algorithm (EG) and an adaptive fusion mechanism (AF) are proposed to further improve the performance of the module. The details are shown in Fig.~\ref{Fig_RG_EG}. Before the prediction head of the branch, we introduce the region-aware spatial propagation network (RASPN) based on the constraint of the semantic prior for further refinement.

\subsection{Dense Repetitive Hourglass Network}\label{DRHN}
In challenging environments for autonomous driving, understanding the contexts of color images becomes particularly critical, given the extreme sparsity of depth measurements. The problem of blurred image guidance can be mitigated by using a powerful feature extractor that can obtain clear contextual semantics. Our innovative dense repetitive hourglass network, which builds on existing backbones \cite{He2016Deep,tan2019efficientnet} with a certain depth, focuses on expanding the width \cite{xue2022go}. 

For a single DRHN$_i$: As illustrated in Fig.~\ref{Fig_pipeline}, the original color image is first encoded by a $3\times3$ convolution and then fed into DRHN$_1$. In the encoder of DRHN$_i$, $\mathbf E_{ij}^I$ takes ${\mathbf E}_{i(j-1)}^I$ and $\mathbf D_{(i-1)j}^I$ as input, while in the decoder of DRHN$_i$, $\mathbf D_{ij}^I$ inputs $\mathbf E_{ij}^I$ and $\mathbf D _{i(j+1)}^I$. The process can be formulated as: 
\begin{equation}\label{e1}
\begin{split}
& {{\mathbf E}_{ij}^I}=f^c({{\mathbf E}_{i(j-1)}^I})+{{\mathbf D}_{(i-1)j}^I},\, \ \ 1<i, \ j\le 5,\\
& {{\mathbf D}_{ij}^I}=f^{d}({{\mathbf D}_{i(j+1)}^I})+{{\mathbf E}_{ij}^I},\ \ \ \ \ \ \ \ 1\le j<5. 
\end{split}
\end{equation}
where $f^c(\cdot)$ and $f^{d}(\cdot)$ denote convolution and deconvolution functions, respectively. In addition, $\mathbf E_{1j}^I=f^c(\mathbf E_{1(j-1)}^I)$, $\mathbf E_{i1}^I=f^c({{\mathbf D}_{(i-1)1}^I})$, and $\mathbf D_{i5}^I=f^c({{\mathbf E}_{i5}^I})$. 

For multiple DRHNs: After deploying DRHN$_1$, we stack the similar symmetric units repeatedly to gradually extract discriminative features. To lower the complexity of DRHNs, we perform only two convolutions in each layers of DRHN$_i$ ($i>1$). Moreover, following the idea of DenseNet \cite{huang2017densely}, we construct DRHN by establishing dense connections among different repetitive hourglass units. In other words, as depicted in Fig.~\ref{Fig_DRHN}, 
the input of DRHN$_i$ is derived from the outputs of DRHN$_1$, DRHN$_2$, $\cdots$, and DRHN$_{i-1}$, which are mapped by a non-linear transformation $f^{t}(\cdot)$ that comprises batch normalization (BN) \cite{ioffe2015batch}, rectified linear unit (ReLU) \cite{glorot2011deep}, and convolution. We formulate this process as: 
\begin{equation}\label{e_dense_connection}
{\mathbf E}_{ij}^I=f^c({{\mathbf E}_{i(j-1)}^I})+f^t(\sum\nolimits_{s=1}^{i-1}{{\mathbf D}_{sj}^I}), \ \ \ 1<i\le 5. 
\end{equation}
Different from DenseNet \cite{huang2017densely}, which densely connects multiple stages within a single encoder, our DRHN treats each hourglass network as an independent unit that receives the outputs of all previous hourglass networks. 

\subsection{Semantic Prior Guidance}\label{semantic_guidance_branch}
In addition to the color image guidance, we leverage the prior knowledge of the large-scale vision model SAM \cite{kirillov2023segment}, which generates explicit high-level semantic cues that can also offer coarse structure guidance through its mask edges. 

As illustrated in Fig.~\ref{Fig_pipeline}, the semantic guidance branch first applies a $3\times 3$ convolution to encode the semantic map input. Then, a symmetric UNet-like network, which has the same architecture as DRHN$_1$, is employed. We define this process as follows: 
\begin{equation}\label{e_semantic_branch}
\begin{split}
& {{\mathbf E}_{1j}^S}=f^c({{\mathbf E}_{1(j-1)}^S}),\ \ \ \ \ \ \quad \, \, \, \ \ 1<j\le5,\\
& {{\mathbf D}_{1j}^S}=f^{d}({{\mathbf D}_{1(j+1)}^S})+{{\mathbf E}_{1j}^S}, \ \ \ 1\le j<5. 
\end{split}
\end{equation}

\subsection{Repetitive Guidance Module}\label{RG}
Depth in challenging environments is not only extremely sparse but also diverse. Most of the existing methods suffer from unclear structures, especially near the object boundaries. Following previous gradual refinement methods \cite{Cheng2020CSPN,park2020nonlocal} that are proven effective, we propose the repetitive guidance module (see Fig.~\ref{Fig_RG_EG}) to progressively generate dense and structure-detailed depth. 

As shown in Fig.~\ref{Fig_pipeline}, our depth generation branch has the same architecture as DRHN$_1$. Given the image guidance feature $\mathbf D_{ij}^I$, semantic guidance feature $\mathbf D_{1j}^S$, and sparse depth feature $\mathbf E_{1j}^{Sp}$, our repetitive guidance (RG) module employs the efficient guidance algorithm (see Sec.~\ref{AG}) to produce refined depth $\mathbf d_{jk}$ step by step. Then, RG fuses the refined $\mathbf d_{jk}$ by the adaptive fusion mechanism (see Sec.~\ref{AF}) to obtain the depth feature $\mathbf d_j$: 
\begin{equation}\label{e2}
{{\mathbf d}_{j}}=f^{RG}({{\mathbf D}_{ij}^I}, \ {{\mathbf D}_{1j}^S}, \ {{\mathbf E}_{1j}^{Sp}}),
\end{equation}
where $f^{RG}(\cdot)$ refers to the repetitive guidance function. As a result, the depth network is described as:
\begin{equation}\label{e_depth_branch}
\begin{split}
& {{\mathbf E}_{1j}^{Sp}}=f^c({{\mathbf d}_{j-1}}), \ \ \ \quad \quad \quad \, \, \, \ \ 1<j<5,\\
& {{\mathbf D}_{1j}^{Sp}}=f^{d}({{\mathbf D}_{1(j+1)}^{Sp}})+{{\mathbf E}_{1j}^{Sp}},\ \ \ 1\le j<5.
\end{split}
\end{equation}
At the last layer of the encoder ($j=5$), ${{\mathbf E}_{15}^{Sp}}=f^c({{\mathbf d}_{4}})+{{\mathbf D}_{i5}^{I}}$. 

\subsubsection{Efficient Guidance Algorithm}\label{AG}
Given $\mathbf D_{ij}^I$, $\mathbf D_{1j}^S$, and $\mathbf E_{1j}^{Sp}$, which have the same size of $C\times H\times W$. $R^2$ is the value of the filter kernel size. Then the theoretical complexity of the dynamic convolution is $O(C\times C\times {{R}^{2}}\times H\times W)$. In general, $C$, $H$, and $W$ are usually very large. It is thus necessary to reduce the complexity of the dynamic convolution. As an alternative, GuideNet \cite{tang2020learning} proposes the channel-wise and cross-channel convolution factorization, whose complexity is $O(C\times {R^2}\times H\times W + C\times C)$. 

However, our RG performs the factorization many times, and the channel-wise process still consumes a significant amount of GPU memory, \emph{i.e.}, $O(C\times {R^2}\times H\times W)$. Inspired by SENet \cite{hu2018squeeze} that captures high-frequency response with channel-wise enhancement, we design an efficient guidance unit to reduce the complexity of the channel-wise convolution while encoding high-frequency components, as shown in Fig.~\ref{Fig_RG_EG}. Specifically, for the channel-wise process, we concatenate the image, semantic, and depth inputs, followed by a $3\times 3$ convolution and a global average pooling function, yielding a $C\times 1\times 1$ filter. We then perform pixel-wise dot between this filter and the depth input. For the cross-channel process, the $C\times 1\times 1$ filter is transformed into a $C\times C$ filter by a $3\times 3$ convolution, which is conducted matrix multiplication along with the output of the channel-wise convolution. 
The complexity of our channel-wise convolution is only $O(C\times H \times W)$, reduced to ${1}/{{{R}^{2}}}$. When $k>1$: 
\begin{equation}\label{e3}
{{\mathbf d}_{jk}}=f^{EG}(f^c({{\mathbf D}_{ij}^I}), \ f^c({{\mathbf D}_{1j}^S}), \ {{\mathbf d}_{k-1}}),
\end{equation}
where $f^{EG}(\cdot)$ represents the efficient guidance function. In the case of $k=1$, ${\mathbf d}_{j1}=f^{EG}({{\mathbf D}_{ij}^I}, \ {{\mathbf D}_{1j}^S}, \ {{\mathbf E}_{1j}^{Sp}})$. 

\begin{table}[t]
\caption{Numerical analysis on GPU memory consumption.}\label{t1}
\centering
\renewcommand\arraystretch{1.1}
\resizebox{0.36\textwidth}{!}{
\begin{tabular}{l|ccc}
\hline
\toprule
Method                  & DC     & CF & EG \\
\midrule
Memory (GB)             & 42.75           & 0.334                 & 0.037  \\
Times (-$/$EG)          & 1,155            & 9                     & 1   \\
\bottomrule
\hline
\end{tabular}
}
\end{table}

Suppose the memory consumption of the common dynamic convolution, convolution factorization, and our EG are $M_{DC}$, $M_{CF}$, and $M_{EG}$, respectively. Then we yield: 
\begin{equation}\label{e_gpu_memory_EG_DC}
\begin{split}
&\frac{{{M}_{EG}}}{{{M}_{DC}}}=\frac{C\times H\times W+C\times C}{C\times C\times {{R}^{2}}\times H\times W}=\frac{H\times W+C}{C\times H\times W\times {{R}^{2}}}, \\
&\frac{{{M}_{EG}}}{{{M}_{CF}}}=\frac{C\times H\times W+C\times C}{C\times {{R}^{2}}\times H\times W+C\times C}=\frac{H\times W+C}{C+H\times W\times {{R}^{2}}}.
\end{split}
\end{equation}

Eq.~\ref{e_gpu_memory_EG_DC} shows the theoretical analysis of GPU memory consumption. Under the setting of the second fusion stage (4 in total) in our depth generation branch, using 4-byte floating precision and taking $C=128$, $H=128$, $W=608$, and $R=3$. Tab.~\ref{t1} reports that, the GPU memory of EG is significantly reduced from $42.75$GB to $0.037$GB compared with the common dynamic convolution, nearly $1,155$ times lower in one fusion stage. Besides, comparing to the convolution factorization proposed in \cite{tang2020learning}, the memory of EG is reduced from $0.334$GB to $0.037$GB, nearly $9$ times lower. Therefore, we can conduct our repetitive strategy easily without worrying much about GPU memory consumption. 

\subsubsection{Adaptive Fusion Mechanism}\label{AF}
Given that our RG generates numerous coarse depth features ($\mathbf d_{j1}$, \ ${\mathbf d}_{j2}$, \ $\cdots$, \ $\mathbf d_{jk}$), it is intuitive to employ them collectively for the creation of refined depth maps. This approach has demonstrated its effectiveness across a range of related methodologies \cite{zhao2017pyramid,lin2017feature,Cheng2020CSPN,song2020channel,park2020nonlocal,hu2020PENet}. Inspired by the selective kernel convolution in SKNet \cite{li2019selective}, we propose the adaptive fusion mechanism to refine depth, as shown in Fig.~\ref{Fig_RG_EG}. Specifically, given the coarse depth features, we first concatenate them and then perform a $3\times 3$ convolution. Next, the global average pooling is employed to produce a $C\times 1\times 1$ feature map. Then another $3\times 3$ convolution and a softmax function are applied, obtaining $(\alpha_{1}, \ \alpha_{2}, \ \cdots, \ \alpha_{k})$: 
\begin{equation}\label{e6}
{{\alpha}_{k}}=\sigma(f^{cgc}({\mathbf d}_{j1}, \ {\mathbf d}_{j2}, \ \cdots, \ {\mathbf d}_{jk}),
\end{equation}
where $\sigma(\cdot)$ denotes softmax function whilst $f^{cgc}(\cdot)$ refers to the function that is comprised of a $3\times 3$ convolution, global average pooling, and another $3\times 3$ convolution. As a result, we fuse $k$ coarse depth maps using $\alpha_{k}$ to produce $\mathbf d_j$: 
\begin{equation}\label{e7}
{{\mathbf d}_{j}}=\sum\nolimits_{r=1}^{k}{{{\alpha }_{r}}{{\mathbf d}_{jr}}}.
\end{equation}
Overall, Eqs.~\ref{e6} and~\ref{e7} can be combined as: 
\begin{equation}\label{e8}
{{\mathbf d}_{j}}=f^{AF}({\mathbf d}_{j1}, \ {\mathbf d}_{j2}, \ \cdots, \ {\mathbf d}_{jk}),
\end{equation}
where $f^{AF}(\cdot)$ represents the adaptive fusion function.

Last but not least, to further decrease the complexity of our repetitive guidance module, we replace all of its $3\times 3$ convolutions with depth-wise separable convolutions \cite{howard2017mobilenets}.

 \begin{figure}[t]
  \centering
  \includegraphics[width=1\columnwidth]{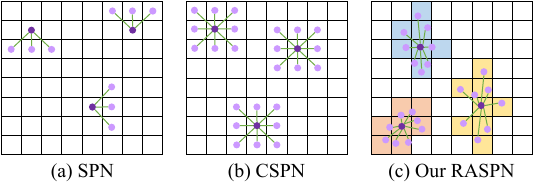}\\
  \caption{Comparison of SPNs, including SPN \cite{liu2017learning}, CSPN \cite{Cheng2020CSPN}, and our RASPN. The colored regions in (c) refer to the object classes provided by semantic masks.}\label{Fig_RASPN}
\end{figure}

\subsection{Region-Aware Spatial Propagation Network}\label{RASPN}
For the depth completion task, SPN methods \cite{cspneccv,park2020nonlocal,lin2022dynamic,liu2022graphcspn} are widely used to recursively produce refined depth results. For simplicity, we replace ${\mathbf D}_{11}^{Sp}$ with ${\mathbf X}$, which will be updated by spatial propagation. Let ${\mathbf X}_{a,b}$ denotes the pixel at $(a,b)$. ${\mathbf N}_{a,b}$ refers to its neighbors, one of which is located at $(m,n)$, \emph{i.e.}, ${{(m,n)}\in {\mathbf N}_{a,b}}$. The propagation of ${\mathbf X}_{a,b}$ at step $(t+1)$ is defined as:
\begin{equation}\label{eq_spn_propagation}
\mathbf{X}_{a,b}^{t+1}=(1-\sum\limits_{(m,n)}{\omega _{a,b}^{m,n}})\mathbf{X}_{a,b}^{t}+\sum\limits_{(m,n)}{\omega _{a,b}^{m,n}\mathbf{X}_{m,n}^{t}},
\end{equation}
where $\omega _{a,b}^{m,n}$ indicates the affinity between the pixels at $(a,b)$ and $(m,n)$. 

As depicted in Fig.~\ref{Fig_RASPN}, the fundamental concept of SPNs lies in the manner of identifying the reasonable and accurate neighbor set ${\mathbf N}_{a,b}$. The original SPN \cite{liu2017learning} in Fig.~\ref{Fig_RASPN} (a) presents a three-way local connection to build $\mathbf{N}_{a,b}^{spn}$, where each pixel is linked to three adjacent pixels of one row or column. The left-to-right propagation is described as:
\begin{equation}\label{eq_raw_spn}
\mathbf{N}_{a,b}^{spn}=\{\mathbf{X}_{a+u,b+v} \ | \ u\in \{-1,0,1 \}, \ v=1 \}.
\end{equation}
The other directions, \emph{i.e.}, right-to-left, top-to-bottom, and bottom-to-right, can be defined in similar ways. 
Moreover, CSPN \cite{cspneccv} in Fig.~\ref{Fig_RASPN} (b) constructs $\mathbf{N}_{a,b}^{cspn}$ within a fixed square area excluding the centre pixel: 
\begin{equation}\label{eq_cspn}
\mathbf{N}_{a,b}^{cspn}=\{\mathbf{X}_{a+u,b+v} \ | \ u,v\in \{-1,0,1 \}, \ (u,v)\neq (0,0) \}.
\end{equation}

Different from SPN and CSPN, our RASPN, as shown in Fig.~\ref{Fig_RASPN} (c), incorporates the semantic knowledge of $\mathbf{D}_{11}^S$, thereby providing explicit object masks. The propagation of $\mathbf{X}_{a,b}$ is specifically confined to its mask region $\mathbf{M}_{a,b}$: 
\begin{equation}\label{eq_raspn}
\mathbf{N}_{a,b}^{cspn}=\{\mathbf{X}_{a+u,b+v} \ | \ (u,v)\in \mathbf{M}_{a,b} \}.
\end{equation}
This explicit mask constraint facilitates the learning of a more accurate affinity, particularly in the vicinity of object edges. In addition, we employ the deformable convolutions \cite{zhu2019deformable,park2020nonlocal} to enhance the non-local propagation within $\mathbf{M}_{a,b}$.

\subsection{Loss Function}\label{loss_func} 
Since ground-truth depth is often semi-dense in real world, following \cite{tang2020learning,zhang2023cf}, we compute the loss only at the valid pixels of the ground-truth depth. Given the final refined depth $\mathbf X_{p}$ in RASPN, and the corresponding ground-truth depth $\mathbf Y_{p}$, we define the reconstruction loss as: 
\begin{equation}\label{eq_loss}
L_{recons}=\frac{1}{{|\mathbb{P}}|}\sum\limits_{p\in {{\mathbb{P}}}}{\left( \mathbf Y_{p}-{\mathbf X}_{p} \right)}^{2},
\end{equation}
where $\mathbb{P}$ is the set of valid pixels of $\mathbf Y_{p}$ and $p$ is one pixel of it. $|\mathbb{P}|$ denotes the total number of the valid pixels. 

\section{TOFDC Dataset}

\subsection{Motivation}
For depth completion, KITTI \cite{Uhrig2017THREEDV} and NYUv2 \cite{silberman2012indoor} are the two commonly used datasets. Tab.~\ref{tab_dataset_comparison} lists their detailed characteristics. 
KITTI uses LiDAR to collect outdoor scenes, while NYUv2 employs Kinect with time-of-flight (TOF) to capture indoor scenes. However, both LiDAR and Kinect are bulky and inconvenient, especially for ordinary consumers in daily life. Recently, TOF depth sensors have become more common on edge devices (\emph{e.g.}, mobile phones), as depth information is vital for human-computer interaction, such as virtual reality and augmented reality. Therefore, it is important and worthwhile to create a new depth completion dataset on consumer-level edge devices. 

\subsection{Data Collection}
\begin{figure}[t]
  \centering
  \includegraphics[width=1\columnwidth]{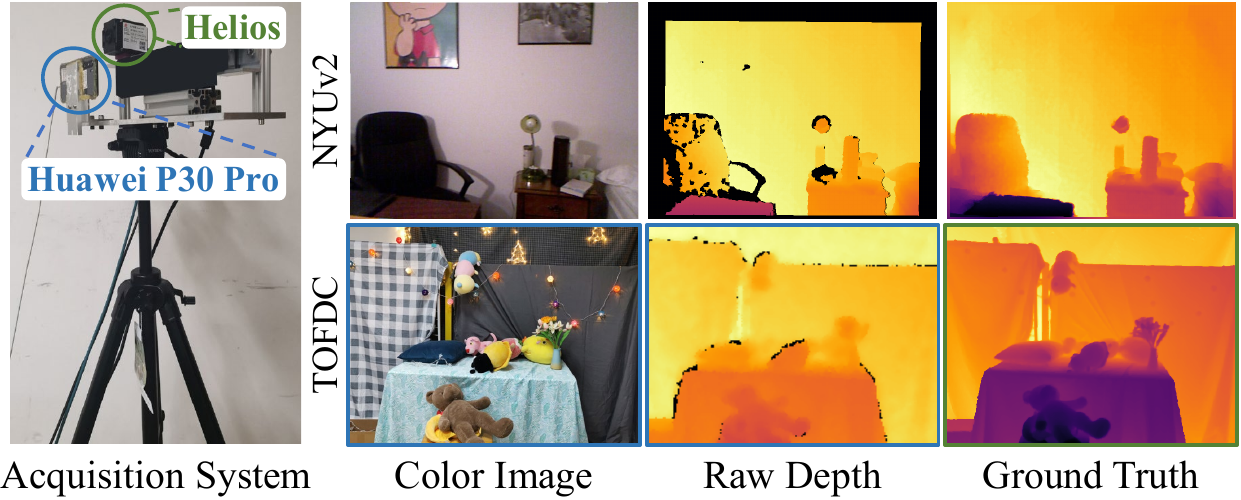}\\
  \caption{Acquisition system and data comparison.}\label{Fig_collection_raw_data}
\end{figure}

\textbf{Acquisition System.} As illustrated in Fig.~\ref{Fig_collection_raw_data} (left), the acquisition system consists of the Huawei P30 Pro and Helios, which capture color image and raw depth, and ground truth depth, respectively. The color camera of P30 produces $3,648\times 2,736$ images using a 40 megapixel Quad Bayer RYYB sensor, while the TOF camera outputs $240\times 180$ raw depth maps. The industrial-level Helios TOF camera generates higher-resolution depth. Their depth acquisition principle is the same, ensuring consistent depth values.

\begin{table*}[t]
\centering
\renewcommand\arraystretch{1.1}
\resizebox{0.95\textwidth}{!}{
\begin{tabular}{l|cc|c|c|ccc|c}
\toprule
Dataset     & Outdoor & Indoor  & Sensor  & Edge Device  & Train  & Test  & Resolution  & Real-world  \\ 
\midrule
KITTI \cite{Uhrig2017THREEDV}    & \checkmark & $\times$    & LiDAR       & $\times$  & 86,898  & 1,000  & $1216\times 352$  & \checkmark  \\
NYUv2 \cite{silberman2012indoor} &  $\times$  & \checkmark  & Kinect TOF  & $\times$  & 47,584  & 654    & $304\times 228$  & $\times$  \\
TOFDC                            & \checkmark & \checkmark  & Phone TOF   & \checkmark  & 10,000  & 560    & $512\times 384$  & \checkmark  \\
\bottomrule
\end{tabular}
}
\caption{Dataset comparison. Note that these characteristics are calculated according to the \emph{depth completion task}.}\label{tab_dataset_comparison}
\end{table*}

 \begin{figure}[t]
  \centering
  \includegraphics[width=1\columnwidth]{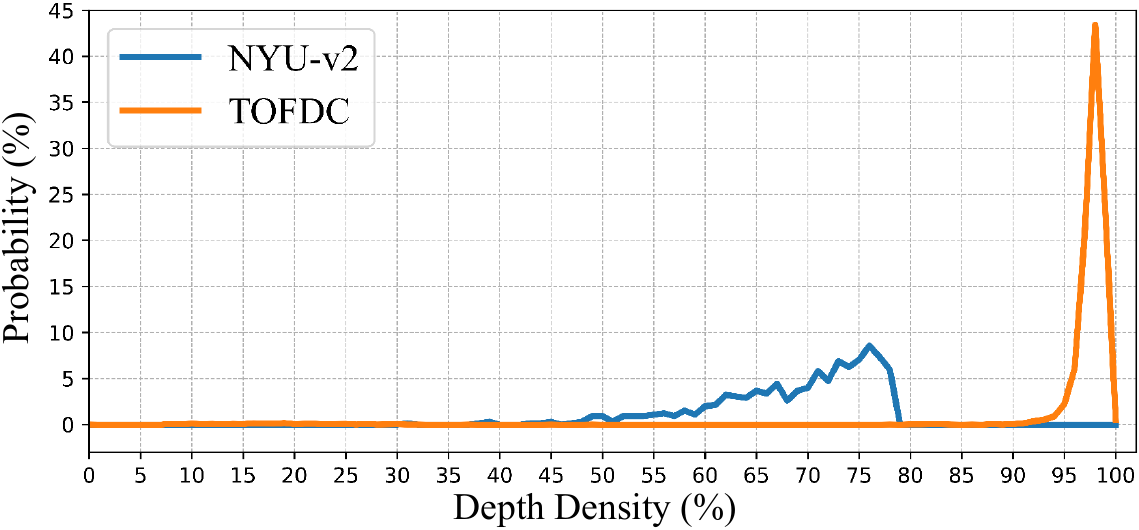}\\
  \caption{Density-probability comparison of raw depth maps.}\label{Fig_tofdc_statistic_density_prob}
\end{figure}

\textbf{Data Processing.} We calibrate the RGB-D system of the P30 with the Helios TOF camera. We align them on the $640\times 480$ color image coordinate using the intrinsic and extrinsic parameters. The color images and Helios depth maps are cropped to $512\times 384$, while the P30 depth maps to $192\times 144$. Then we conduct nearest interpolation to upsample the P30 depth maps to $512\times 384$. For the Helios depth maps, there still exist some depth holes caused by environment and object materials (\emph{e.g.}, transparent glass). We use the colorization technique \cite{levin2004colorization} to fill the holes. 

\begin{figure}[t]
  \centering
  \includegraphics[width=0.92\columnwidth]{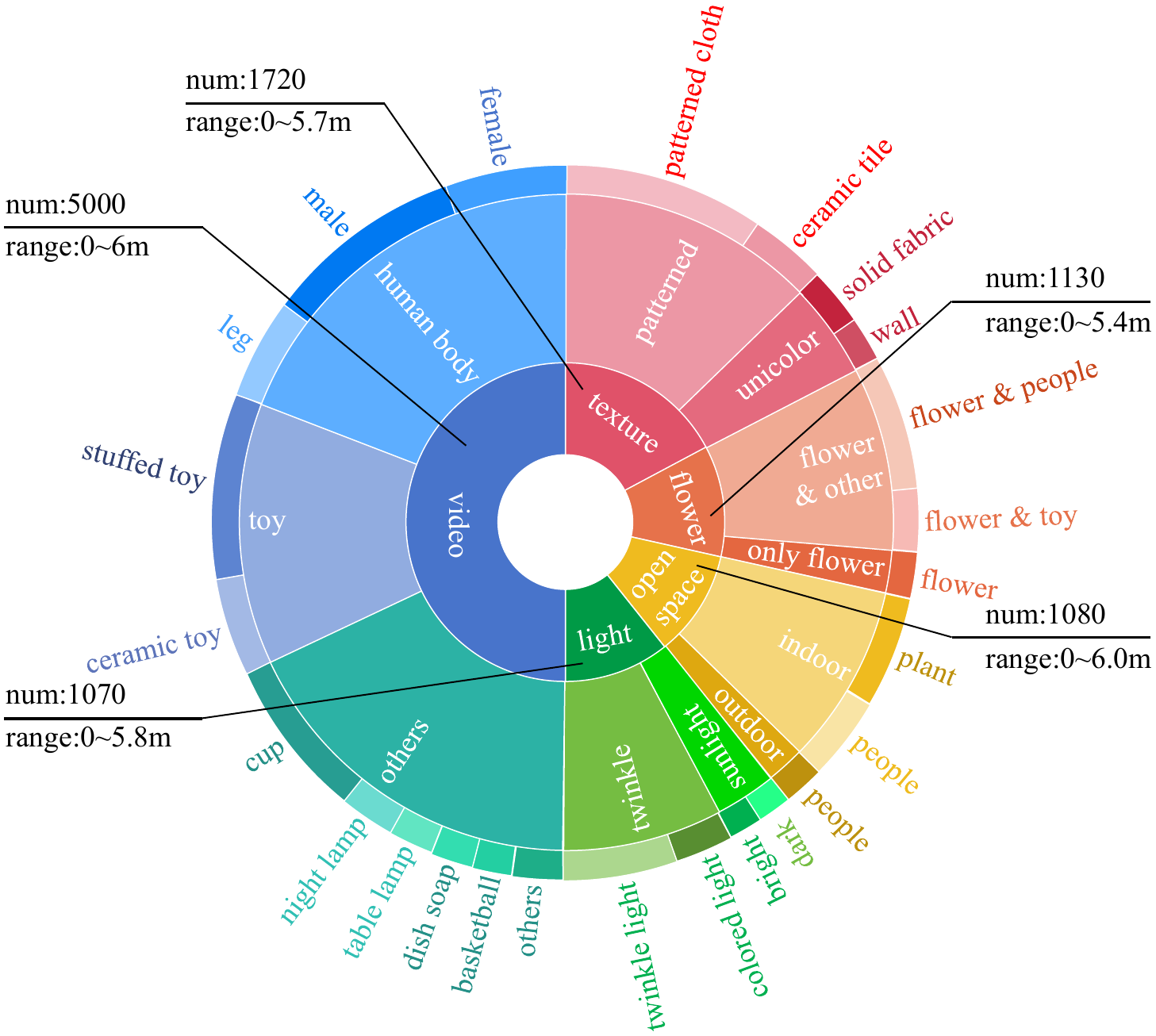}\\
  \caption{Distribution of different scenarios in our TOFDC.
  }\label{Fig_tofdc_scene_statistic}
\end{figure}

\textbf{Results.}
Based on the acquisition system and data processing, we have collected the new depth completion dataset TOFDC. As reported in Tab.~\ref{tab_dataset_comparison}, it consists of both indoor and outdoor scenes, including texture, flower, light, video, and open space, 10,000 RGB-D pairs in total. For the depth completion task, we take the raw depth captured by the P30 TOF lens as input, which is different from NYUv2 where the input depth is sampled from the ground truths. Fig.~\ref{Fig_collection_raw_data} (right) shows the post-processing depth results. We can observe that the depth of TOFDC is much denser than that of NYUv2. Fig.~\ref{Fig_tofdc_statistic_density_prob} provides the corresponding statistical support. It reveals that the depth density of NYUv2 varies mainly from 60\% to 80\%, whereas that of TOFDC is highly concentrated between 95\% and 100\%. Fig.~\ref{Fig_tofdc_scene_statistic} shows the detailed distribution of the five scenarios of TOFDC.

\section{Experiments}
In this section, we first introduce datasets, metrics, and implementation details. Then, we carry out extensive experiments to evaluate our method against other state-of-the-art works. Finally, a number of ablation studies are employed to verify the effectiveness of our RigNet++.

\subsection{Datasets and Metrics} 
\textbf{KITTI Depth Completion Benchmark} \cite{Uhrig2017THREEDV} is a large real-world benchmark for self-driving. The sparse depth is produced by projecting raw LiDAR points through the view of camera. The semi-dense ground-truth depth is generated by first projecting the accumulated LiDAR scans of multiple timestamps, and then removing abnormal depth values from occlusion and moving objects. As a result, the training split has 86,898 ground-truth annotations with aligned sparse LiDAR maps and color images. The official validation split and test split are both comprised of 1,000 pairs. In addition, since there are almost no LiDAR points at the top of depth maps, the input images are bottom center cropped \cite{vangansbeke2019,tang2020learning,zhao2021adaptive,liu2021fcfr} from $1216\times 352$ to $1216\times 256$. 

\textbf{NYUv2} \cite{silberman2012indoor} is comprised of video sequences from a variety of indoor scenes, which are recorded by RGB-D cameras using Microsoft Kinect. Paired color images and depth maps in 464 indoor scenes are commonly used. Following previous depth completion methods \cite{chen2019learning,Qiu_2019_CVPR,park2020nonlocal,tang2020learning}, we train our model on 50K images from the official training split, and test on the 654 images from the official labeled test set. Each image is downsized to $320\times 240$, and then $304\times 228$ center-cropping is applied. As the input resolution of our network must be a multiple of 32, we further pad the images to $320\times 256$, but evaluate only at the valid region of size $304\times 228$ to keep fair comparison with other methods. 

\textbf{TOFDC} is collected by the TOF sensor and RGB camera of a Huawei P30 Pro, which covers various scenes such as texture, flower, body, and toy, under different lighting conditions and in open space. It has 10,000 $512\times 384$ RGB-D pairs for training and 560 for evaluation. The ground truth depth maps are captured by the Helios TOF camera~\footnote{https://thinklucid.com/product/helios-time-of-flight-imx556/}. 

\textbf{Matterport3D} \cite{albanis2021pano3d} is a 360$^\circ$ scanned dataset collected by Matterport's Pro 3D panoramic camera. The latest Matterport3D \footnote{https://vcl3d.github.io/Pano3D/download/} ($512\times 256$) contains 7,907 panoramic RGB-D pairs, of which 5636 for training, 744 for validating, and 1527 for testing. For panoramic depth completion, M$^3$PT \cite{yan2022multi} proposes to synthesize the sparse depth. It first projects the equirectangular ground-truth depth into cubical map to remove the distortion. Next, it generates cube sparse depth via imitating the laser scanning, \emph{e.g.}, taking one pixel for every eight pixels horizontally and one pixel for every two pixels vertically. Then the cube sparse depth is binarized to obtain cube binary mask, which is back projected into spherical plane. Finally, it multiplies the spherical mask by the ground-truth depth to produce the final sparse depth. 

\textbf{3D60} \cite{zioulis2019spherical} is also a panoramic dataset collected from real world. The latest 3D60 \footnote{https://vcl3d.github.io/3D60/} ($512\times 256$) consists of 6,669 RGB-D pairs for training, 906 for validating, and 1831 for testing, 9,406 in total. In the same manner, M$^3$PT provides sparse depth for panoramic depth completion evaluation. 

\textbf{Virtual KITTI} \cite{gaidon2016virtual} is a
synthetic dataset cloned from the
real-world KITTI video sequences. Besides, it produces color images under various lighting (\emph{e.g.}, sunset, morning) and weather (\emph{e.g.}, rain, fog) conditions. Following \cite{tang2020learning}, we use the masks generated from sparse depth of KITTI dataset to obtain sparse samples. Such strategy makes it close to real-world situation for the sparse depth distribution. Sequences of 0001, 0002, 0006, and 0018 are used for training, 0020 with various lighting and weather conditions is used for testing. As a result, it contributes to 1,289 frames for fine-tuning and 837 frames for evaluating each condition.

\begin{table}[t]
\caption{Definition of the seven relevant evaluation metrics.}\label{tab_metric}
\centering
\Large
\renewcommand\arraystretch{1.5}
\resizebox{0.4835\textwidth}{!}{
\begin{tabular}{ll}
\toprule
\multicolumn{2}{l}{For one pixel $p$ in the valid pixel set $\mathbb{P}$: } \\ \midrule
-- REL               & $\frac{1}{|\mathbb{P}|}\sum\limits{{{\left| \mathbf Y_{p}-{\mathbf X}_{p} \right|/{\mathbf Y_{p}}}}}$  \\ 
-- MAE               & $\frac{1}{|\mathbb{P}|}\sum\limits{{{\left| \mathbf Y_{p}-{\mathbf X}_{p} \right|}}}$  \\ 
-- iMAE              & $\frac{1}{|\mathbb{P}|}\sum\limits{{{\left| 1/{\mathbf Y_{p}}-1/{{\mathbf X}_{p}} \right|}}}$  \\ 
-- RMSE              & $\sqrt{\frac{1}{|\mathbb{P}|}\sum\limits{{{\left( \mathbf Y_{p}-{\mathbf X}_{p} \right)}}}^2}$  \\  
-- iRMSE             & $\sqrt{\frac{1}{|\mathbb{P}|}\sum\limits{{{\left( 1/{\mathbf Y_{p}}-1/{{\mathbf X}_{p}} \right)}}}^2}$  \\  
-- RMSELog           & $\sqrt{\frac{1}{|\mathbb{P}|}\sum{{{\left(  {\log {\mathbf Y_p}}-\log {\mathbf X_p} \right)}^{2}}}}$  \\  
-- ${\delta }_{i}$   & $\frac{|\mathbb{S}|}{|\mathbb{P}|}\times 100\%, \ \mathbb{S}: \max \left( {{\mathbf Y_p}/{\mathbf X_p},{\mathbf X_p}/{\mathbf Y_p}} \right)<{1.25}^{i}$  \\ 
\bottomrule
\end{tabular}
}
\end{table}

\textbf{Metrics.} For the outdoor KITTI dataset, following the KITTI benchmark and existing methods \cite{park2020nonlocal,tang2020learning,liu2021fcfr,hu2020PENet}, we use four standard metrics for evaluation, including RMSE, MAE, iRMSE, and iMAE. For the indoor NYUv2 dataset, following previous works \cite{chen2019learning,Qiu_2019_CVPR,park2020nonlocal,tang2020learning,liu2021fcfr}, three metrics are selected, including RMSE, REL, and ${{\delta }_{i}}$ ($i=1, 2, 3$). For the indoor panoramic Matterport3D and 3D60 datasets, following M$^3$PT\cite{yan2022learning}, we employ RMSE, MAE, REL, RMSELog, and ${{\delta }_{i}}$ for test. Please see Tab.~\ref{tab_metric} for more details. 

\subsection{Implementation Details}
The model is trained end-to-end from scratch and implemented on Pytorch framework with 4 TITAN RTX GPUs. We train it for 15 epochs in total employing the loss function that is defined in Eq.~\ref{eq_loss}. We use ADAM \cite{kingma2014adam} as the optimizer with the momentum of $\beta_{1}=0.9$, $\beta_{2}=0.999$, and weight decay of $1 \times {10}^{-6}$. The starting learning rate is $1 \times {10}^{-3}$, which drops by half every 5 epochs. We leverage color jittering and horizontal flip for data augmentation. In addition, the synchronized cross-GPU batch normalization \cite{ioffe2015batch,zhang2018context} is conducted to calculate more accurate 
mean and variance. As a result, the batch size is set to 8. 

\subsection{Evaluation on Outdoor KITTI}
Tab.~\ref{tab_kitti} shows the quantitative comparisons on the KITTI depth completion benchmark, ranked by RMSE. The top part lists the results of 2D-3D joint methods, which introduce normal, graph, point cloud, and other 3D representations. The bottom part reports those of 2D-based approaches that employ semantic, boundary, and other 2D information.

 \begin{figure*}[t]
  \centering
  \includegraphics[width=2.04\columnwidth]{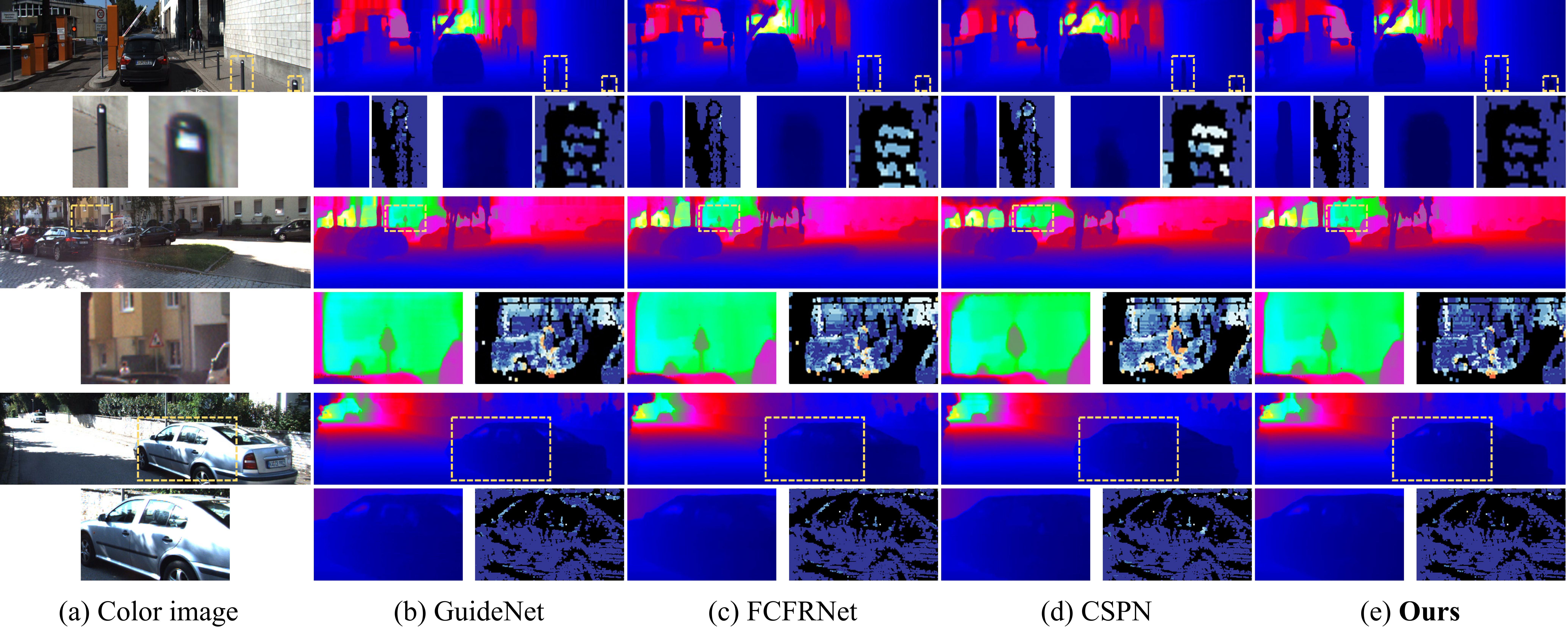}\\
  \caption{Qualitative results on KITTI benchmark, including GuideNet \cite{tang2020learning}, FCFRNet \cite{liu2021fcfr}, CSPN \cite{cspneccv}, and our method. \textbf{Error maps} borrowed from KITTI leaderboard are for detailed discrimination, where warmer color refers to higher error.}\label{Fig_kitti_vis}
\end{figure*}

\begin{table*}[!ht]
\caption{Quantitative results on \href{https://www.cvlibs.net/datasets/kitti/eval_depth.php?benchmark=depth_completion}{KITTI leaderboard}. 2D and 3D indicate that models employ 2D and 3D representations, respectively. $^\star$ denotes models that involve additional training data. The \textbf{best} and the \textcolor{blue}{second best} metrics are highlighted.}\label{tab_kitti}
\centering
\large
\renewcommand\arraystretch{1.1}
\resizebox{1\textwidth}{!}{
\begin{tabular}{l|cc|c|cccc|c}
\toprule
{Method}                           & 2D        & 3D  & Params. (M) $\downarrow$ & RMSE (mm) $\downarrow$     & MAE (mm) $\downarrow$ & iRMSE (1/km) $\downarrow$  & iMAE (1/km) $\downarrow$  & Publication \\ 
\midrule
DepthNormal \cite{Xu2019Depth} & \checkmark & \checkmark & 40.0  & 777.05    & 235.17   & 2.42           & 1.13           & ICCV 2019 \\
FuseNet \cite{chen2019learning}    & \checkmark & \checkmark & 1.9  & 752.88    & 221.19   & 2.34           & 1.14           & ICCV 2019 \\
DLiDAR$^\star$ \cite{Qiu_2019_CVPR}        & \checkmark & \checkmark& 53.4   & 758.38    & 226.50   & 2.56           & 1.15           & CVPR 2019 \\
ACMNet \cite{zhao2021adaptive}     & \checkmark & \checkmark & 4.9 & 744.91    & 206.09   & 2.08           & 0.90           & TIP 2021 \\
PointFusion\cite{huynh2021boosting}& \checkmark & \checkmark & 8.7 & 741.90    & 201.10   & 1.97           & 0.85           & ICCV 2021 \\
GraphCSPN \cite{liu2022graphcspn}  & \checkmark & \checkmark & 26.4 & 738.41    & 199.31   & 1.96           & 0.84           & ECCV 2022 \\
BEV@DC \cite{zhou2023bev}           & \checkmark & \checkmark & 30.8 & 697.44    & \textcolor{blue}{189.44}   & \textcolor{blue}{1.83}           & \textcolor{blue}{0.82}           & CVPR 2023 \\
PointDC \cite{yu2023aggregating}   & \checkmark & \checkmark & 25.1 & 736.07    & 201.87   & 1.97           & 0.87           & ICCV 2023 \\
\midrule
S2D \cite{ma2018self}              & \checkmark  &   & 26.1   & 814.73    & 249.95   & 2.80           & 1.21           & ICRA 2019 \\
NConv \cite{2020Confidence}        & \checkmark  &   & \textbf{0.36}   & 829.98    & 233.26   & 2.60           & 1.03           & PAMI 2020 \\
FusionNet$^\star$ \cite{vangansbeke2019}   & \checkmark  &   & 2.55   & 772.87    & 215.02   & 2.19           & 0.93           & MVA 2019 \\
CSPN++ \cite{Cheng2020CSPN}        & \checkmark  &   & 26.0   & 743.69    & 209.28   & 2.07           & 0.90           & AAAI 2020 \\
NLSPN \cite{park2020nonlocal}      & \checkmark  &   & 25.8   & 741.68    & 199.59   & 1.99           & 0.84           & ECCV 2020 \\
GuideNet \cite{tang2020learning}   & \checkmark  &   & 73.5   & 736.24    & 218.83   & 2.25           & 0.99           & TIP 2020 \\
TWISE \cite{imran2021depth}        & \checkmark  &   & \textcolor{blue}{1.45}   & 840.20    & 195.58   & 2.08           & \textcolor{blue}{0.82}           & CVPR 2021 \\
FCFRNet \cite{liu2021fcfr}         & \checkmark  &   & 50.6   & 735.81    & 217.15   & 2.20           & 0.98           & AAAI 2021 \\
PENet \cite{hu2020PENet}           & \checkmark  &   & 131.5  & 730.08    & 210.55   & 2.17           & 0.94           & ICRA 2021 \\
GFormer \cite{rho2022guideformer}  & \checkmark  &   & 130.0  & 721.48    & 207.76   & 2.14           & 0.97           & CVPR 2022 \\
MFFNet \cite{liu2023mff}   & \checkmark  &  & 122.7  & 719.85  & 208.11  & 2.21  & 0.94  & RAL 2023  \\
DySPN \cite{lin2022dynamic}        & \checkmark  &   & 26.3   & 709.12    & 192.71   & 1.88           & \textcolor{blue}{0.82}           & AAAI 2022 \\ 
CFormer \cite{zhang2023cf}         & \checkmark  &   & 83.5   & 708.87    & 203.45   & 2.01           & 0.88           & CVPR 2023 \\
LRRU \cite{wang2023lrru}           & \checkmark  &   & 21.0   & \textcolor{blue}{696.51}    & 189.96   & 1.87           & \textbf{0.81}  & ICCV 2023 \\ 
\midrule
RigNet \cite{yan2022rignet}        & \checkmark  &   & 65.2   & 712.66    & 203.25   & 2.08           & 0.90           & ECCV 2022 \\
\textbf{RigNet++ (ours)}               & \checkmark  &  & 25.4 & \textbf{694.24} & \textbf{188.62} & \textbf{1.82} & \textbf{0.81} & - \\ 
\bottomrule
\end{tabular}
}
\end{table*}

 \begin{figure*}[t]
  \centering
  \includegraphics[width=2.04\columnwidth]{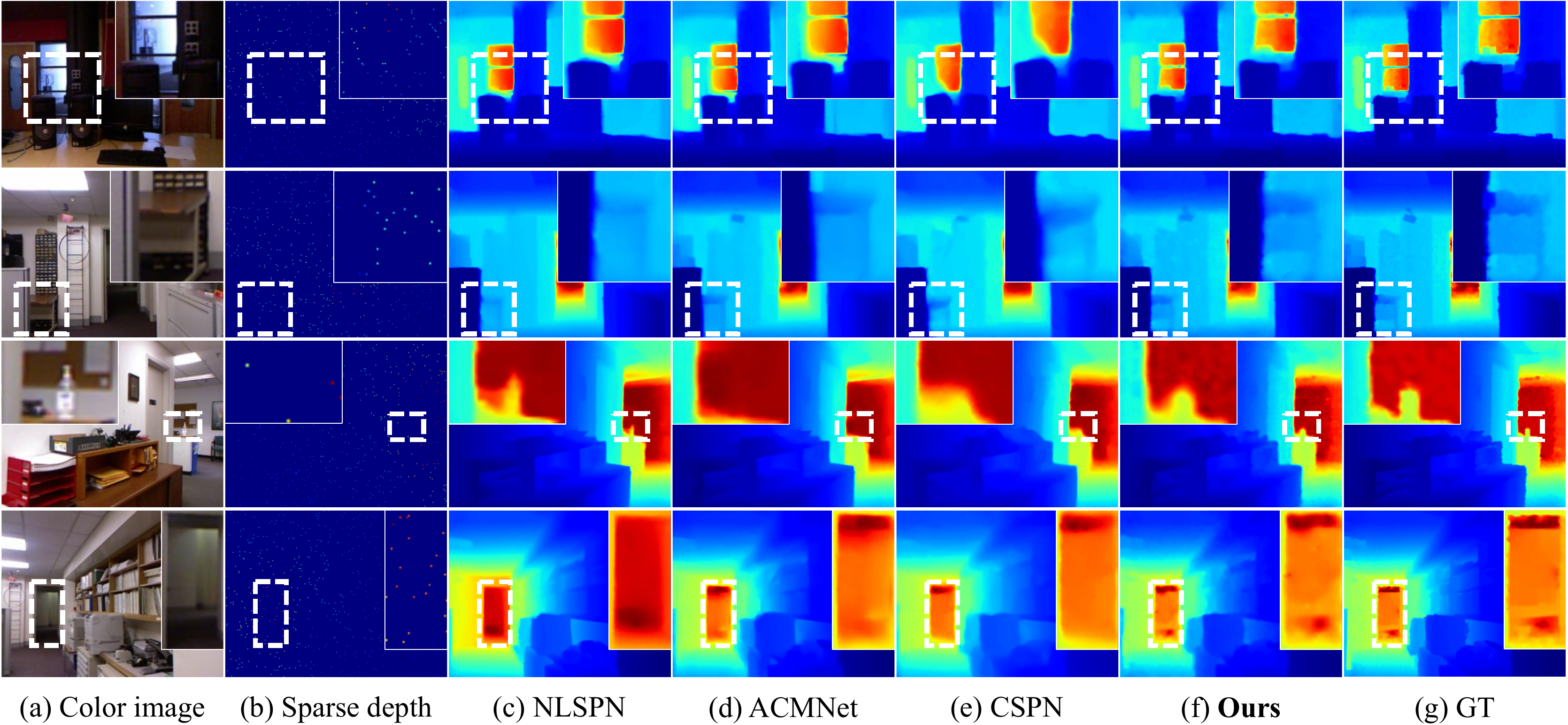}\\
  \caption{Qualitative results on NYUv2 test set. From left to right: (a) color image, (b) sparse depth, (c) NLSPN \cite{park2020nonlocal}, (d) ACMNet \cite{zhao2021adaptive}, (e) CSPN \cite{cspneccv}, (f) our RigNet++, and (g) ground-truth depth. 
  }\label{Fig_nyu_vis}
\end{figure*}

We discover that our RigNet++ outperforms other state-of-the-art methods. For instance, when compared with the 2D-3D joint PointDC \cite{yu20233d}, the RMSE of RigNet++ is lower by 41.83mm. In comparison with the 2D transformer-based models GFormer \cite{rho2022guideformer} and CFormer \cite{rho2022guideformer}, RigNet++ demonstrates superior performance in terms of RMSE, which is 27.24mm and 14.63mm better, respectively. It’s noteworthy that the parameters of GFormer and CFormer are 5.1 and 3.3 times greater than that of RigNet++, severally. 
Besides, our RigNet++ also performs better than these approaches that employ additional datasets. For example, DLiDAR \cite{Qiu_2019_CVPR} is pre-trained on the synthetic dataset produced from CARLA \cite{dosovitskiy2017carla}, jointly learning dense depth and surface normal tasks. Our model surpasses DLiDAR by 64.14mm in RMSE and 37.88mm in MAE. Furthermore, our RigNet++ significantly outperforms those lightweight models, such as TWISE \cite{imran2021depth}, NConv \cite{2020Confidence}, and ACMNet \cite{zhao2021adaptive}. It achieves 110.8mm reduction averagely in the dominant RMSE metric. Finally, it’s important to highlight that our RigNet++ is not only more lightweight but also superior to the earlier version of RigNet. This makes it a more efficient and effective solution. 

Visual comparisons with the state-of-the-art approaches are depicted in Fig.~\ref{Fig_kitti_vis}. Although all methods generally yield visually pleasing results, our model is capable of generating superior depth predictions with greater details and more precise object boundaries. The associated error maps provide further clear visual indicators. For instance, among the highlighted iron pillars, walls, and cars in the 2nd, 4th, and 6th rows of Fig.~\ref{Fig_kitti_vis}, the error in our prediction is significantly less than those of others. 

\begin{table}[!ht]
\caption{Quantitative comparisons on NYUv2 dataset with 500 sparse pixel sampling. The top part refers to 2D-3D joint methods, while the bottom indicates 2D based approaches.}\label{tab_nyuv2}
\centering
\large
\renewcommand\arraystretch{1.1}
\resizebox{0.4835\textwidth}{!}{
\begin{tabular}{l|ccccc}
\toprule
Method      & RMSE (m)      & REL     &${\delta }_{1}$  & ${\delta }_{{2}}$   & ${\delta }_{{3}}$  \\ 
\midrule
Zhang \emph{et al.} \cite{zhang2018deep} & 0.228    & 0.042   & 97.1  & 99.3  & 99.7 \\
DLiDAR \cite{Qiu_2019_CVPR}      & 0.115    & 0.022   & 99.3  & \textbf{99.9}  & \textbf{100.0} \\
Xu \emph{et al.} \cite{Xu2019Depth}  & 0.112  & 0.018  & 99.5  & \textbf{99.9}  & \textbf{100.0}  \\
ACMNet \cite{zhao2021adaptive}   & 0.105    & 0.015   & 99.4  & \textbf{99.9}  & \textbf{100.0} \\
GraphCSPN \cite{liu2022graphcspn}& 0.090    & 0.012   & \textcolor{blue}{99.6}  & \textbf{99.9}  & \textbf{100.0}  \\ 
BEV@DC \cite{zhou2023bev}        & \textcolor{blue}{0.089}    & 0.012   & \textcolor{blue}{99.6}  & \textbf{99.9}  & \textbf{100.0} \\ 
PointDC \cite{yu2023aggregating} & \textcolor{blue}{0.089}    & 0.012   & \textcolor{blue}{99.6}  & \textbf{99.9}  & \textbf{100.0}  \\
\midrule
Bilateral \cite{silberman2012indoor}  & 0.479    & 0.084   & 92.4  & 97.6  & 98.9 \\
S2D\_18 \cite{ma2018sparse}  & 0.230    & 0.044   & 97.1  & 99.4  & 99.8 \\
DCoeff \cite{silberman2012indoor}  & 0.118    & 0.013   & 99.4  & \textbf{99.9}  & - \\
CSPN \cite{cspneccv}             & 0.117    & 0.016   & 99.2  & \textbf{99.9}  & \textbf{100.0} \\
FCFRNet \cite{liu2021fcfr}       & 0.106    & 0.015   & 99.5  & \textbf{99.9}  & \textbf{100.0} \\
PRNet \cite{lee2021depth}        & 0.104    & 0.014   & 99.4  & \textbf{99.9}  & \textbf{100.0} \\
GuideNet \cite{tang2020learning} & 0.101    & 0.015   & 99.5  & \textbf{99.9}  & \textbf{100.0} \\
TWISE \cite{imran2021depth}      & 0.097    & 0.013   & \textbf{99.6}  & \textbf{99.9} & \textbf{100.0} \\ 
NLSPN \cite{park2020nonlocal}    & 0.092    & 0.012   & \textcolor{blue}{99.6}  & \textbf{99.9}  & \textbf{100.0} \\ 
DySPN \cite{lin2022dynamic}      & 0.090    & 0.012   & \textcolor{blue}{99.6}  & \textbf{99.9}  & \textbf{100.0} \\
CFormer \cite{zhang2023cf}       & 0.091    & 0.012   & \textcolor{blue}{99.6}  & \textbf{99.9}  & \textbf{100.0} \\
LRRU \cite{wang2023lrru}         & 0.091    & \textcolor{blue}{0.011}   & \textcolor{blue}{99.6}  & \textbf{99.9}  & \textbf{100.0}  \\
\midrule
RigNet                           & 0.090    & 0.013   & \textcolor{blue}{99.6}  & \textbf{99.9}  & \textbf{100.0} \\ 
\textbf{RigNet++ (ours)}         & \textbf{0.087}    & \textbf{0.010}   & \textbf{99.7}  & \textbf{99.9}  & \textbf{100.0}   \\ 
\bottomrule
\end{tabular}
}
\end{table}

 \begin{figure*}[t]
  \centering
  \includegraphics[width=2.04\columnwidth]{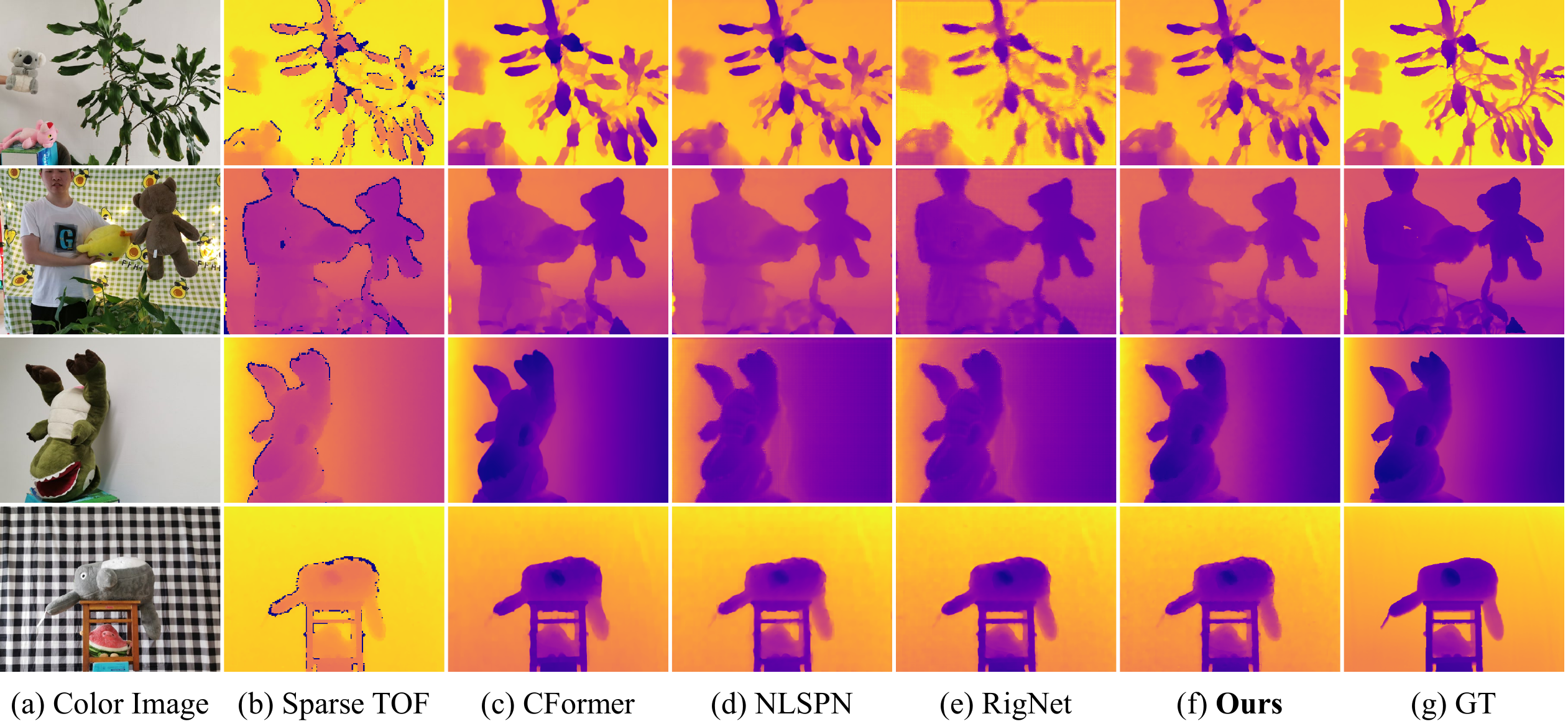}\\
  \caption{Qualitative results on TOFDC dataset. From left to right: (a) color image, (b) sparse TOF, (c) CFormer \cite{zhang2023cf}, (d) NLSPN \cite{park2020nonlocal}, (e) RigNet \cite{yan2022rignet}, (f) our RigNet++, and (g) ground-truth depth.}\label{Fig_tofdc_vis}
\end{figure*}

\begin{table}[t]
\caption{Quantitative comparisons on our TOFDC dataset.}\label{tab_tofdc}
\centering
\large
\renewcommand\arraystretch{1.1}
\resizebox{0.4835\textwidth}{!}{
\begin{tabular}{l|ccccc}
\toprule
Method          & RMSE (m)       & REL      &${\delta }_{1}$   & ${\delta }_{{2}}$  & ${\delta }_{{3}}$   \\ 
\midrule 
GraphCSPN \cite{liu2022graphcspn} & 0.253  & 0.052  & 92.0  & 96.9  & 98.7  \\ 
PointDC \cite{yu2023aggregating}  & \textcolor{blue}{0.109}  & \textcolor{blue}{0.021}  & \textcolor{blue}{98.5}  & 99.2  & 99.6  \\
\midrule 
CSPN \cite{cspneccv}              & 0.224  & 0.042  & 94.5  & 95.3  & 96.5  \\
FusionNet \cite{vangansbeke2019}  & 0.116  & 0.024  & 98.3  & \textcolor{blue}{99.4}  & \textcolor{blue}{99.7}  \\
GuideNet \cite{tang2020learning}  & 0.146  & 0.030  & 97.6  & 98.9  & 99.5  \\
ENet \cite{hu2020PENet}           & 0.231  & 0.061  & 94.3  & 95.2  & 97.4  \\
PENet \cite{hu2020PENet}          & 0.241  & 0.043  & 94.6  & 95.3  & 95.5  \\
NLSPN \cite{park2020nonlocal}     & 0.174  & 0.029  & 96.4  & 97.9  & 98.9  \\
CFormer \cite{zhang2023cf}        & 0.113  & 0.029  & \textbf{99.1}   & \textbf{99.6}  & \textbf{99.9}  \\
\midrule
RigNet \cite{yan2022rignet}       & 0.133  & 0.025  & 97.6  & 99.1  & \textcolor{blue}{99.7}  \\
\textbf{RigNet++ (ours)}              & \textbf{0.097}    & \textbf{0.016}   & \textbf{99.2}  & \textbf{99.6}  & \textbf{99.9}  \\ 
\bottomrule
\end{tabular}
}
\end{table}

\subsection{Evaluation on Indoor NYUv2}
To verify the performance of RigNet++ on indoor scenes, following existing approaches \cite{Cheng2020CSPN,park2020nonlocal,tang2020learning,liu2021fcfr}, we train it on NYUv2 dataset under the setting of 500 sparse depth pixel sampling. The sparse example can be found in Fig.~\ref{Fig_nyu_vis}(b). 

As listed in Tab.~\ref{tab_nyuv2}, our model outperforms all previous and recent methods. For example, RigNet++ surpasses the 2D-based methods by an average of 56.0mm in RMSE and is 31.3mm superior to the 2D-3D joint approaches. Even when compared to its earlier version, RigNet \cite{yan2022rignet}, RigNet++ still exhibits better performance across all five metrics. 

Fig.~\ref{Fig_nyu_vis} demonstrates the qualitative visualization results. Obviously, compared with those state-of-the-art methods, our model can recover more detailed structures with lower errors at most pixels, including sharper boundaries and more complete object shapes. For example, among the marked doors in the last row of Fig.~\ref{Fig_nyu_vis}, our prediction is very close to the ground-truth annotation, while others either have large errors in the whole regions or have blurry shapes on specific objects. These facts give strong evidence that the proposed method manages to work in indoor scenes.

\subsection{Evaluation on Indoor \& Outdoor TOFDC}
To further evaluate our RigNet++, we have implemented it on the newly proposed TOFDC dataset, which is collected using consumptive TOF sensors. All the methods listed in Tab.~\ref{tab_tofdc} are retrained from scratch using available codes. 

The results indicate that our RigNet++ significantly outperforms the 2D-3D joint approaches. For instance, it reduces the RMSE by 61.7\% and REL by 69.2\% compared to GraphCSPN \cite{liu2022graphcspn}, which incorporates point cloud prior. RigNet++ also demonstrates a 12mm advantage over the second best method, PointDC \cite{yu2023aggregating}, in terms of RMSE. Furthermore, when compared with the best 2D based method CFormer \cite{zhang2023cf}, our RigNet++ shows outstanding improvement of 16mm in RMSE, which is a significant enhancement for indoor scenes. Fig.~\ref{Fig_tofdc_vis} reveals that RigNet++ is capable of estimating high-quality dense depth results with clearer and sharper structures, \emph{e.g.}, stuffed toys, cartons, flowers, human bodies, and textureless regions.

 \begin{figure*}[t]
  \centering
  \includegraphics[width=2.06\columnwidth]{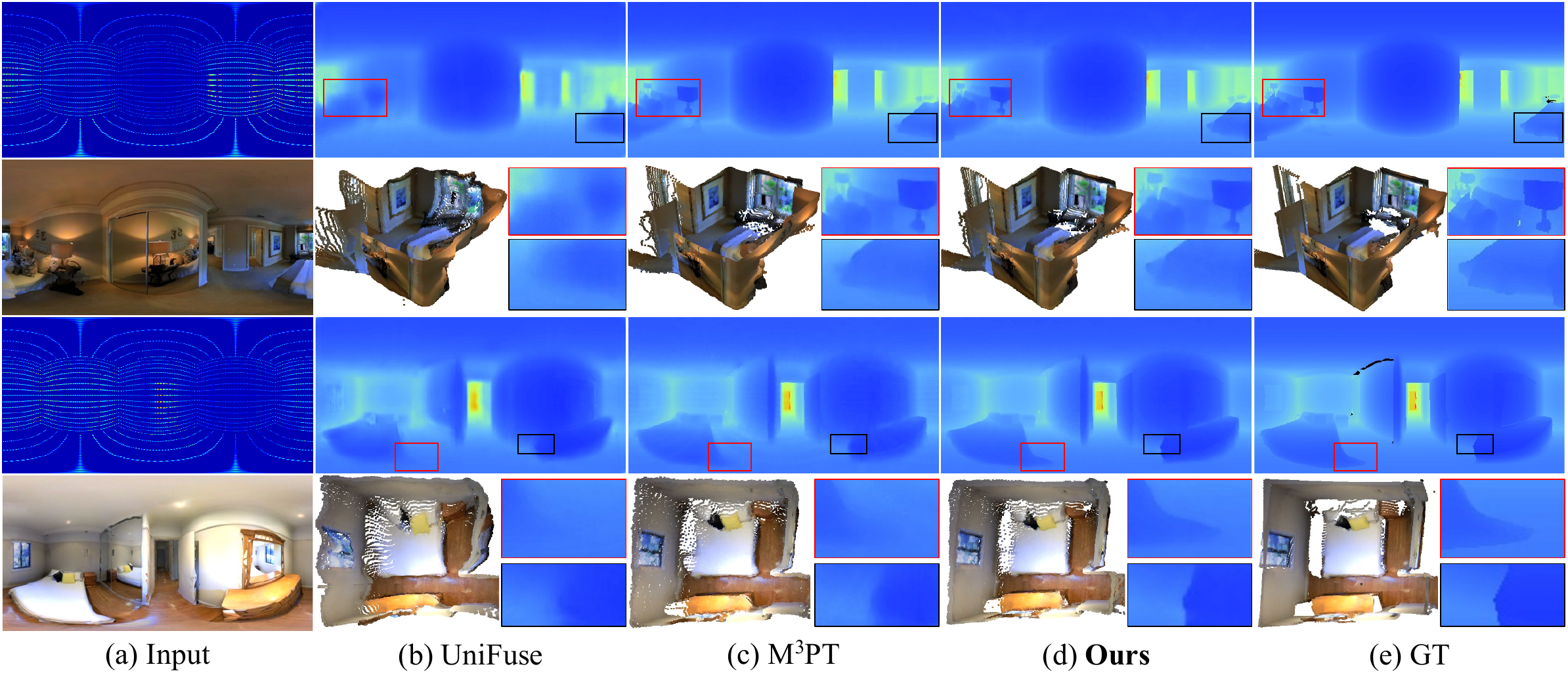}\\
  \caption{Qualitative results on Matterport3D and 3D60 datasets. From left to right: (a) sparse depth and color image input, (b) UniFuse \cite{jiang2021unifuse}, (c) M$^3$PT \cite{yan2022multi}, (d) our method, and (e) ground-truth depth.}\label{Fig_360_vis}
  \vspace{9pt}
\end{figure*}

\begin{table*}[t]
\caption{Quantitative comparisons on panoramic Matterport3D and 3D60 datasets. Results are borrowed from M$^3$PT~\cite{yan2022multi}.} 
\label{tab_360}
\centering
\renewcommand\arraystretch{1.1}
\resizebox{0.995\textwidth}{!}{
\begin{tabular}{c|l|c|ccccccc} 
\toprule
Dataset  & Method  & Params. (M)  & RMSE (mm) & MAE (mm)  & REL & RMSELog  & ${\delta }_{1}$  & ${\delta }_{{2}}$ & ${\delta }_{{3}}$   \\
\midrule
\multirow{8}{*}{\begin{sideways}{Matterport3D}\end{sideways}} 
& UniFuse \cite{jiang2021unifuse}   & \textcolor{blue}{30.3}    & 229.1  & 95.2  & 0.0475  & 0.0381  & 97.10  & 99.24  & 99.70   \\
& HoHoNet \cite{sun2021hohonet}   & 54.1       & 199.2  & 75.0  & 0.0355  & 0.0311  & 98.06  & 99.45  & 99.77  \\
& PENet \cite{hu2020PENet}      & 131.5    & 248.0  & 91.5  & 0.0493  & 0.0350  & 97.28  & 99.35  & 99.70  \\ 
& GuideNet \cite{tang2020learning}  & 73.5    & 192.9  & 87.2  & 0.0438  & 0.0327  & 98.06  & 99.48  & 99.81    \\
& 360Depth~\cite{rey2022360monodepth}  & -  & 185.3  & 68.8  & 0.0302  & 0.0285  & 98.33  & 99.42  & 99.80   \\
& M$^3$PT~\cite{yan2022multi}  & 74.2     & \textcolor{blue}{138.9}  & \textcolor{blue}{36.2}  & \textcolor{blue}{0.0164}  & \textcolor{blue}{0.0193}  & \textcolor{blue}{99.27}  & \textcolor{blue}{99.76}  & \textcolor{blue}{99.90}   \\ \cmidrule{2-10}
& RigNet   & 65.2        & 155.8  & 55.9  & 0.0271  & 0.0245  & 98.84  & 99.66  & 99.86  \\ 
& \textbf{RigNet++} (ours)  & \textbf{25.4}    & \textbf{113.5}  & \textbf{35.2}  & \textbf{0.0155}  & \textbf{0.0176}  & \textbf{99.40}  & \textbf{99.85}  & \textbf{99.95}     \\  \midrule
\multirow{8}{*}{\begin{sideways}{3D60}\end{sideways}}        
& UniFuse \cite{jiang2021unifuse} & \textcolor{blue}{30.3}      & 215.6  & 94.1  & 0.0446  & 0.0342  & 97.49  & 99.47  & 99.84   \\
& HoHoNet \cite{sun2021hohonet}  & 54.1       & 196.9  & 75.6  & 0.0338  & 0.0294  & 98.18  & 99.54  & 99.83   \\
& PENet \cite{hu2020PENet}   & 131.5   & 233.9  & 120.3 & 0.0680  & 0.0321  & 97.43  & 99.26  & 99.80  \\
& GuideNet \cite{hu2020PENet}  & 73.5      & 239.3  & 144.2 & 0.0689  & 0.0418  & 97.11  & 99.54  & 99.87    \\
& 360Depth~\cite{rey2022360monodepth} & -  & 225.4  & 93.7  & 0.0677  & 0.0315  & 97.82  & 99.36  & 99.85   \\
& M$^3$PT~\cite{yan2022multi}  & 74.2   & \textcolor{blue}{127.2}  & \textcolor{blue}{34.1}  & \textcolor{blue}{0.0144}  & \textcolor{blue}{0.0165}  & \textcolor{blue}{99.44}  & \textcolor{blue}{99.85}  & \textcolor{blue}{99.95}   \\ \cmidrule{2-10}
& RigNet  & 65.2      & 128.8  & 43.3  & 0.0188  & 0.0185  & 99.30  & 99.83  & 99.94 \\ 
& \textbf{RigNet++} (ours)  & \textbf{25.4}        & \textbf{102.6}  & \textbf{33.2}  & \textbf{0.0143}  & \textbf{0.0146}  & \textbf{99.62}  & \textbf{99.92}  & \textbf{99.96}       \\ 
\bottomrule
\end{tabular}
\vspace{9pt}
}
\end{table*}

\subsection{Evaluation on Panoramic Matterport3D and 3D60}
The task of panoramic depth completion is initially introduced in the M$^3$PT study \cite{yan2022multi}. Subsequently, we implement our RigNet++ on two panoramic depth completion datasets, namely Matterport3D \cite{albanis2021pano3d} and 3D60 \cite{zioulis2019spherical}. The comparison results are documented in Tab.~\ref{tab_360}. 

We can observe that even with the least parameters, our RigNet++ significantly surpasses UniFuse \cite{jiang2021unifuse}, HoHoNet \cite{sun2021hohonet}, GuideNet \cite{tang2020learning}, 360Depth \cite{rey2022360monodepth}, and M$^3$PT \cite{yan2022multi}. For instance, in comparison to M$^3$PT, which employs an additional masked pre-training strategy \cite{he2022masked}, RigNet++ still achieves a 15.1\% lower average RMSE and higher $\delta_i$, even though their REL metrics are marginally different. Furthermore, thanks to the new module and backbone design, our RigNet++ performs better and is less complex compared to its early version, RigNet \cite{yan2022rignet}. Fig.~\ref{Fig_360_vis} presents a visual comparison, including depth prediction and 3D reconstruction results constructed from the panoramic RGB-D pairs. We find that RigNet++ can predict depth results more accurately than other methods. Even when compared to the two-stage M$^3$PT, RigNet++ is still capable of estimating more detailed object edges and surfaces, for example, beds, tables, windows, and walls. All these results clearly indicate the potential application value of our RigNet++ in tasks related to panoramic scene analysis and understanding. 

To sum up, our method is effective, and more efficient than its early version, as shown by these numerical and visual results. We think our model can benefit various tasks, such as 3D scene reconstruction, AR, VR, self-driving, \emph{etc}.

\begin{table*}[t]
\caption{Ablation studies of RHN on KITTI validation set. \textcolor{blue}{18-1} denotes that we use 1 ResNet-18 as the baseline, where the raw guided convolution in GuideNet \cite{tang2020learning} is employed. `Deeper' (or `More') denotes that we conduct single (or multiple \& tandem) hourglass unit to build the network. Note that each layer of RHN$_{2,3}$ only contains two convolutions while the RHN$_{1}$ employs ResNets. `Ckpt size' refers to the size of model checkpoint. RHN with dense connection is ablated in (I) of Tab.~\ref{t_RG}.}
\label{t_RHN}
\centering
\renewcommand\arraystretch{1.1}
\resizebox{1\textwidth}{!}{
\begin{tabular}{l|cccc|ccc|ccc|cccc}
\toprule
\multirow{2}{*}{Method} 
& \multicolumn{4}{c|}{Deeper} & \multicolumn{3}{c|}{More} & \multicolumn{3}{c|}{Deeper-More} & \multicolumn{4}{c}{Our parallel RHN}   \\ \cline{2-15} 
& 10-1  & \textcolor{blue}{18-1}  & 34-1  & 50-1      
& 18-2  & 18-3  & 18-4     
& 34-2  & 34-3  & 50-2    
& 10-2  & 10-3  & 18-2  & 18-3  \\ \midrule
Params. (M) & 59 & \textcolor{blue}{63}  & 71  &  84  & 72  & 81 & 91  & 89 & 107 &  104  & 60   &  61   & 64  & 65       \\
Ckpt size (M) & 224 & \textcolor{blue}{239}  & 273 & 317 & 274 & 309 & 344 & 339 & 407 & 398 & 228 & 232 & 242 & 246 \\
RMSE (mm)    & 822  & \textcolor{blue}{779}  & 778     & 777   & 802     &  816    & 811 & 807  & 801 & 800 & 803 & 798 & 772     & \textbf{769}    \\ 
\bottomrule
\end{tabular}
}
\end{table*}

 \begin{figure*}[t]
  \centering
  \includegraphics[width=2.06\columnwidth]{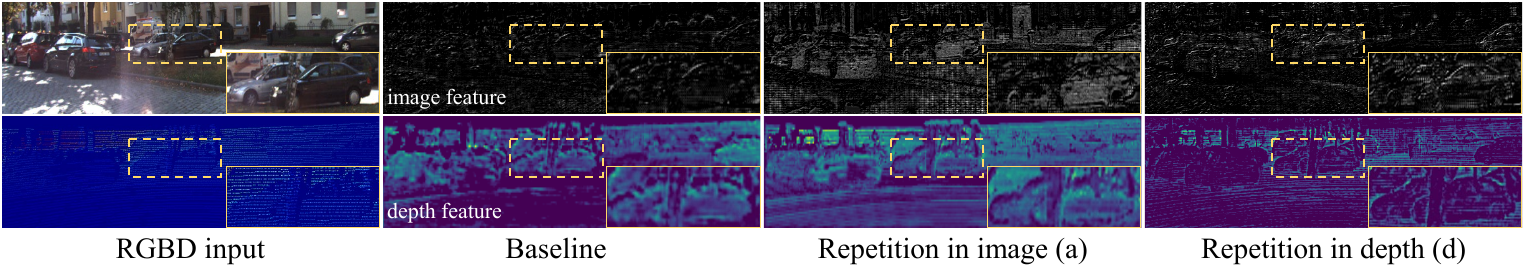}\\
  \caption{Visual comparisons of intermediate features of the baseline and our repetition in image and depth branches.}\label{Fig_intermediate_feature}
\end{figure*}

\subsection{Ablation Studies} 
Here we ablate on KITTI validation split to verify the effectiveness of the dense repetitive hourglass network (DRHN) and repetitive guidance module (RG) in Tabs.~\ref{t_RHN} and~\ref{t_RG}, respectively. RG consists of the efficient guidance algorithm (EG) and adaptive fusion mechanism (AF). Detailed difference of intermediate features produced by our repetitive design is shown in Fig.~\ref{Fig_intermediate_feature}. In Tab.~\ref{t_RG}, the batch size is set to 8 when computing the GPU memory consumption.

\noindent \textbf{(1) Effect of Repetitive Hourglass Network}
The baseline GuideNet \cite{tang2020learning} uses a single ResNet-18 and the guided convolution G$_1$ to predict dense depth. To validate the effect of RHN, we explore the backbone design of the image guidance branch from four different aspects, as listed in Table~\ref{t_RHN}. It is worth noting that the semantic guidance branch in not involved when ablating RHN.

\textbf{(i) Deeper single backbone vs. RHN.} The `Deeper' column of Tab.~\ref{t_RHN} shows that, when replacing the single ResNet-10 with ResNet-18, the error is reduced by 43mm. However, when deepening the baseline from 18 to 34/50, the errors have barely changed, indicating that simply increasing the network depth of image guidance branch cannot bring much benefit for the depth completion task. Differently, with little sacrifice of parameters ($\pm$2 M), our RHN-10-3 and RHN-18-3 are 24mm and 10mm superior to Deeper-10-1 and Deeper-18-1, respectively. Fig.~\ref{Fig_intermediate_feature} demonstrates that the image feature of our parallel RHN-18-3 has much clearer and richer contexts than that of the baseline Deeper-18-1.

\textbf{(ii) More tandem backbones vs. RHN.} As shown in `More' column of Tab.~\ref{t_RHN}, we stack the hourglass unit in series. The models of More-18-2, More-18-3, and More-18-4 have worse performance than the baseline Deeper-18-1. It turns out that the combination of tandem hourglass units is not sufficient to provide clearer image semantic guidance for the depth recovery. In contrast, our parallel RHN achieves better results with fewer parameters and smaller model sizes. These facts give strong evidence that the parallel repetitive design in image guidance branch is effective.

 \begin{figure}[t]
  \centering
  \includegraphics[width=0.94\columnwidth]{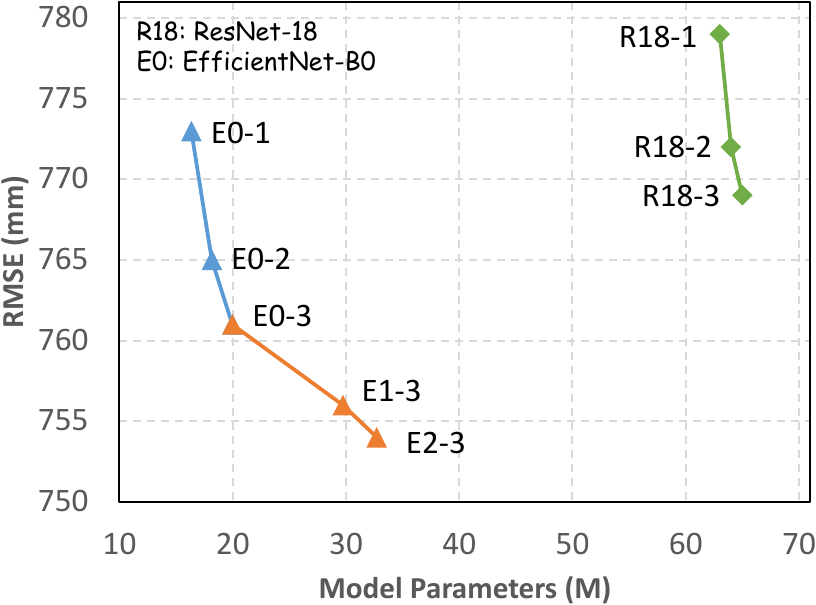}\\
  \caption{Ablation studies of RHN with different backbones.}\label{Fig_ab_RHN}
\end{figure}

\begin{table*}[t]
\caption{Ablation studies of RG, AF, and dense connection strategies on KITTI official validation set. RG-EG$_k$ means that we repeatedly use EG $k$ times. `$\pm0$' refers to 23.37GB. G$_1$ represents the raw guided convolution in GuideNet \cite{tang2020learning}.} 
\label{t_RG}
\centering
\renewcommand\arraystretch{1.1}
\resizebox{0.96\textwidth}{!}{
\begin{tabular}{l|cc|ccccc|ccc|c|c|cc}
\toprule
\multirow{2}{*}{Method}  & \multicolumn{2}{c|}{RHN$_{3}$}  & \multicolumn{5}{c|}{RG} & \multicolumn{3}{c|}{AF}  & Dense  & \multirow{2}{*}{RASPN}  & Memory        & RMSE  \\ \cline{2-11}
& R18  & E0  & G$_1$ & EG$_1$ & EG$_2$ & EG$_3$ & seg & add  & concat & ours  & connection  & & (GB)    & (mm) \\ \midrule
baseline  &  &  &\checkmark &&&& &&&&  & & $\pm$0  & 778.6 \\ \midrule
(a)  & \checkmark  &  &\checkmark &&& &&&&&  & & +1.35  &  769.0 \\ 
(b)         &\checkmark          &              & & \checkmark  &    &         &            &            &              &           & & & -10.60   & 768.6 \\
(c)         &\checkmark        &                      &          &   & \checkmark &    &         &            &              &           & & & +2.65    & 762.3 \\
(d)          &\checkmark      &                       &          &  &  & \checkmark & &            &              &           & & & +13.22   & 757.4 \\
(e)          &\checkmark      &                       &          &  &  & \checkmark & \checkmark &            &              &           & & & +14.57   & 752.8 
\\ 
\midrule
(f)           & \checkmark    &                       &          &  &  & \checkmark & & \checkmark &              &           & & & +13.22   & 755.8 \\
(g)          &\checkmark      &                       &          &  & & \checkmark &  &            & \checkmark   &           & & & +13.22    & 754.6 \\
(h)          & \checkmark     &                        &          &  &  & \checkmark &  &            &              &\checkmark  & & & +13.28 & 752.1 \\ \midrule
(i)          &      & \checkmark                       &          &  &  & \checkmark &  &            &              &\checkmark  & & & +1.06 & 748.2 \\ 
(j)          &      & \checkmark                       &          &  &  & \checkmark &  &            &              &\checkmark  & \checkmark & & +1.27 & 743.5 \\
(k) &      & \checkmark                       &          &  &  & \checkmark &  &            &              &\checkmark  & \checkmark & \checkmark & +2.93 & 734.7 \\
(l) &      & \checkmark                       &          &  &  & \checkmark & \checkmark  &            &              &\checkmark  & \checkmark & \checkmark & +2.94 & \textbf{728.2} \\
\bottomrule
\end{tabular}
}
\end{table*}

\textbf{(iii) Deeper-More backbones vs. RHN.} As reported in `Deeper-More' column of Tab.~\ref{t_RHN}, deeper hourglass units are deployed in serial way. We discover that such combinations are also not very effective, since the errors of them are higher than the baseline while RHN's error is 10mm lower. It verifies the effectiveness of the lightweight RHN design again. 

\textbf{(iv) Different backbones in RHN.} Fig.~\ref{Fig_ab_RHN} shows the results of RHN using different backbones, \emph{i.e.}, ResNet \cite{He2016Deep} and EfficientNet \cite{tan2019efficientnet}. Under the setting of our parallel repetitive design, the errors decrease and the parameters increase when widening (\emph{e.g.}, R18-1/2/3 and E0-1/2/3) or deepening (\emph{e.g.}, E0/1/2-3) the model. Besides, while replacing ResNet with EfficientNet, the errors are further dropped with much fewer parameters. For example, E0-3 is 8mm superior to R18-3 in RMSE whilst the parameters are about 47M lower. What's more, E1-3 and E2-3 can reduce errors with acceptable parameter increasements, which however are still more lightweight than R18-3. We believe that the efficiency and effectiveness of our model can be further improved with other excellent backbones \cite{tan2021efficientnetv2,li2022efficientformer,li2023rethinking}.

\noindent \textbf{(2) Effect of Repetitive Guidance Module}

\textbf{(i) Efficient guidance.} Note that we directly output the features in EG$_{3}$ when not employing AF. Tab.~\ref{t1} has provided quantitative analysis in theory for EG design. Based on (a), we disable G$_1$ by replacing it with EG$_{1}$. Comparing (b) with (a) in Tab.~\ref{t_RG}, both of which carry out the guided convolution technology only once, although the error of (b) goes down a little bit, the GPU memory is considerably reduced by 11.95GB. These results give strong evidence that our new guidance design is not only effective but also efficient. 

\textbf{(ii) Repetitive guidance.} When the recursion number $k$ of EG increases, the errors of (c) and (d) are 6.3mm and 11.2mm much lower than that of (b), respectively. Meanwhile, as shown in Fig.~\ref{Fig_intermediate_feature}, since our repetition in depth (d) can continuously model high-frequency components, the intermediate depth feature has more detailed boundaries while the image guidance branch has a consistently high response nearby the regions. These facts forcefully demonstrate the effectiveness of our repetitive guidance design. 

 \begin{figure}[t]
  \centering
  \includegraphics[width=0.955\columnwidth]{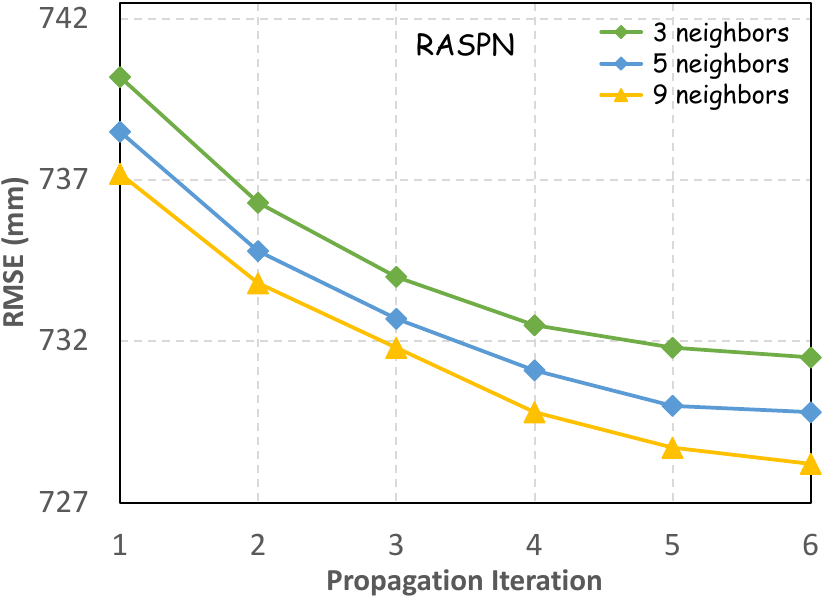}\\
  \caption{Ablation studies of RASPN with different numbers of affinitive neighbors and propagation iterations.}\label{Fig_ab_RASPN}
\end{figure}

 \begin{figure*}[t]
  \centering
  \includegraphics[width=2.06\columnwidth]{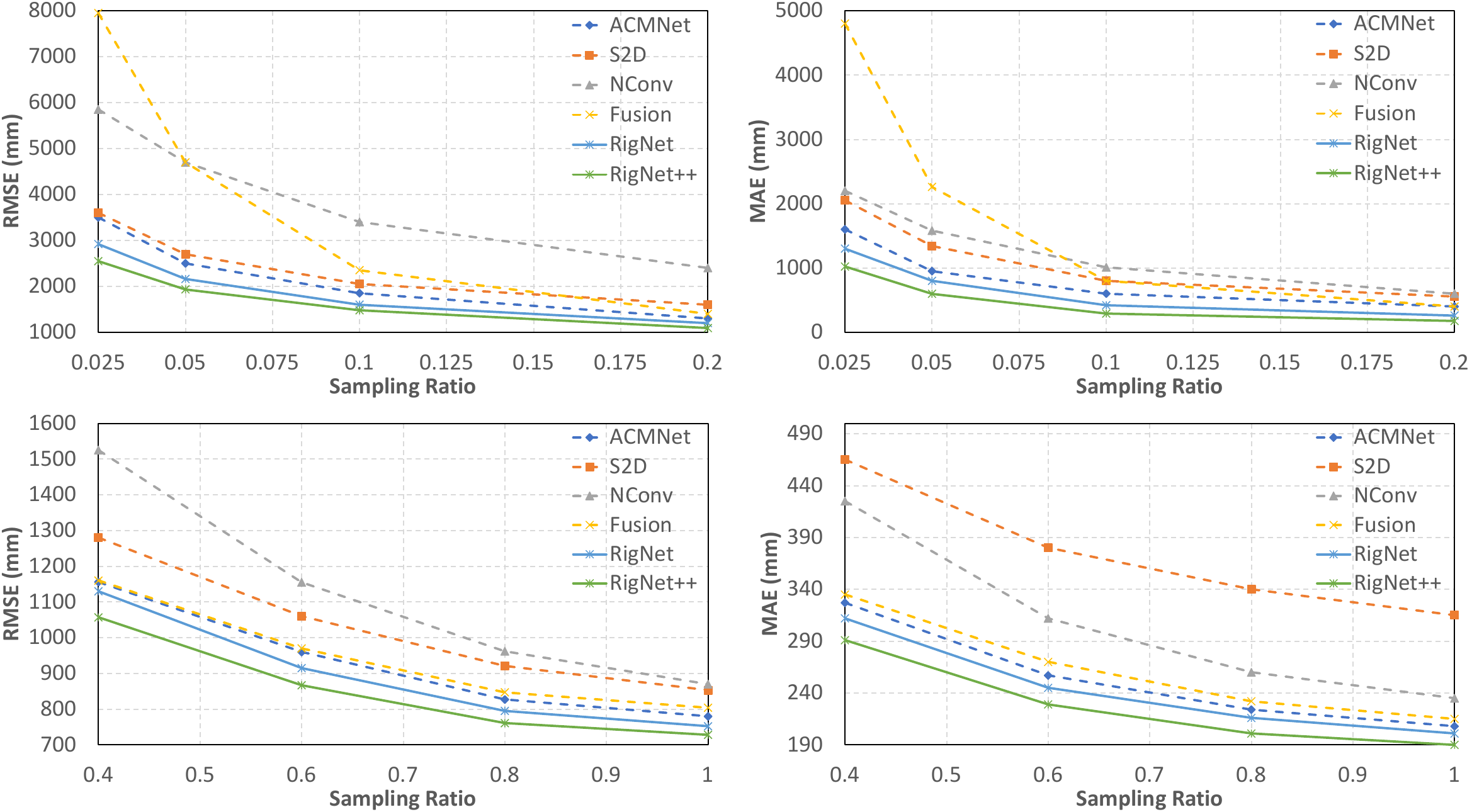}\\
  \caption{Comparisons under different sparsity levels on KITTI validation split, where the solid lines refer to our approaches.}\label{Fig_density_kitti}
\end{figure*}

\textbf{(iii) Adaptive fusion.} Based on (d) that directly outputs the feature of RG-EG$_3$, we leverage all features of RG-EG$_k$ ($k=1,2,3$) to produce better depth representations. (f), (g), and (h) refer to addition, concatenation, and our AF strategies, respectively. Specifically in (g), we conduct a $3 \times 3$ convolution to control the channel to be the same as that of RG-EG$_3$ after concatenation. From `AF' column of Tab.~\ref{t_RG} we discover that, all of the three strategies improve the performance of the model with a little bit GPU memory sacrifice (about 0-0.06GB), which demonstrates that aggregating multi-step features in repetitive procedure is effective. Furthermore, our AF mechanism obtains the best result among these three fusion manners, outperforming (d) 5.3mm. These facts prove that our AF design is more beneficial to the system than those simple fusion strategies. 

\textbf{(iv) Efficient backbone.} When comparing (i) with (h), which substitutes ResNet-18 with EfficientNet-B0, the error decreases by 3.9mm, while the GPU memory is significantly reduced by 12.22GB. Furthermore, when implementing our dense connection strategy across RHNs, the RMSE is further reduced by 4.7mm, with only a minor increase in GPU memory usage of 0.21GB. These quantitative results reaffirm the effectiveness and efficiency of our design. 

\textbf{(3) Effect of Semantic Prior} 

Compared to (d), (e) incorporates the semantic prior into RG, resulting in a reduction of the RMSE from 757.4mm to 752.8mm. Moreover, (k) employs RASPN based on (j), leading to an additional decrease in error by 8.8mm. These outcomes underscore the efficacy of the semantic prior. In Fig.~\ref{Fig_ab_RASPN}, it delves deeper into the influence of affinitive neighbors and propagation iterations in RASPN, revealing that an increased number of neighbors and iterations enhance performance. This suggests that a larger quantity of neighbors and iterations aid in generating more precise local adjacent knowledge. Lastly, (l) integrates all the proposed modules, exceeding the baseline by 50.4mm in RMSE, while only requiring an additional memory consumption of 2.94GB.

 \begin{figure*}[t]
  \centering
  \includegraphics[width=2.02\columnwidth]{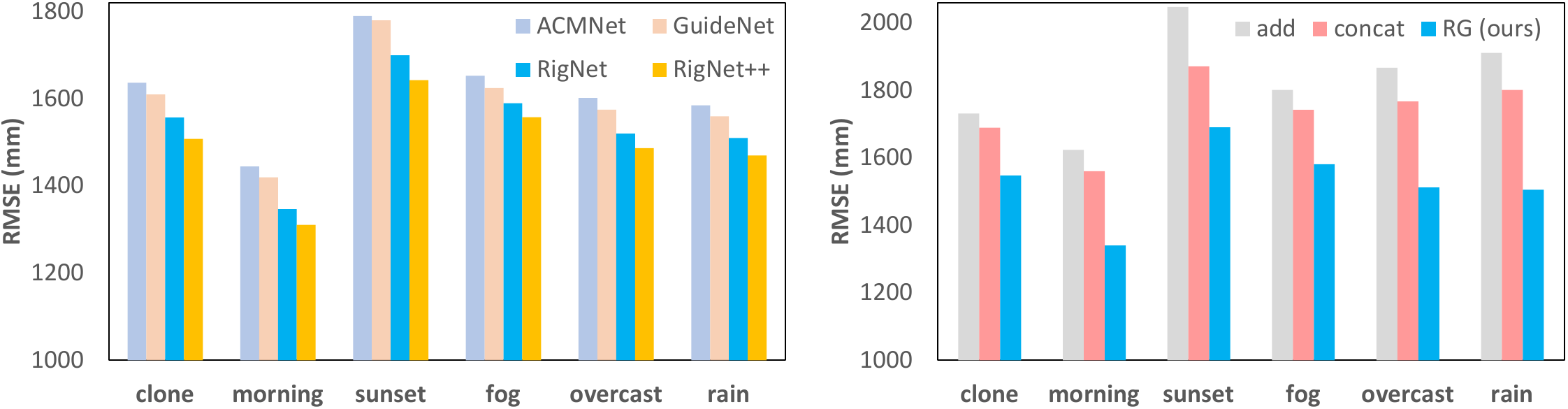}\\
  \caption{Comparisons with existing methods (left) and itself (right) replacing `RG' with `add', under different lighting and weather conditions on Virtual KITTI test split.}\label{Fig_Lighting_and_Weather}
\end{figure*}

\subsection{Generalization Capabilities}
This subsection further validates the generalization capabilities of our models, including the number of valid points in Fig.~\ref{Fig_density_kitti}, various lighting and weather conditions in Fig.~\ref{Fig_Lighting_and_Weather}, and the synthetic pattern of sparse depth in Tab.~\ref{density_nyu}.

\noindent \textbf{(1) Number of valid points}

On KITTI validation split, we compare our method with four well-known approaches with available codes, \emph{i.e.}, S2D \cite{ma2018self}, Fusion \cite{vangansbeke2019}, NConv \cite{2020Confidence}, and ACMNet \cite{zhao2021adaptive}. Note that, all models are pretrained on KITTI training split with raw sparsity, which is equivalent to the sampling ratio of 1.0, but not fine-tuned on the generated sparse depth. Specifically, we first uniformly sample the raw depth maps with ratios (0.025, 0.05, 0.1, 0.2) and (0.4, 0.6, 0.8, 1.0) to produce the sparse depth inputs. Then we test the pretrained models on these inputs. Fig. \ref{Fig_density_kitti} shows that our RigNet++ significantly outperforms others under all levels of sparsity in terms of both RMSE and MAE. These results indicates that our method can deal well with complex data inputs. 

\noindent \textbf{(2) Lighting and weather condition}

The lighting condition of KITTI dataset is almost invariable and the weather condition is good. However, both lighting and weather conditions are vitally important for depth completion, especially for self-driving service. Therefore, we fine-tune our models (trained on KITTI) on `clone' of Virtual KITTI \cite{gaidon2016virtual} and test under all other different lighting and weather conditions. As shown in the right of Fig.~\ref{Fig_Lighting_and_Weather}, we compare `RG' with `add' and `concat' (replace RG with addition and concatenation), our method surpasses `add' and `concat' with large margins in RMSE. The left of Fig. \ref{Fig_Lighting_and_Weather} further reveals that RigNet++ performs better than GuideNet \cite{tang2020learning} and ACMNet \cite{zhao2021adaptive}, and its early version, RigNet \cite{yan2022rignet} in complex environments, whilst RigNet++ achieves best performance. These results demonstrate that our method is able to handle polytropic lighting and weather conditions.

 \begin{figure}[t]
  \centering
  \includegraphics[width=0.9\columnwidth]{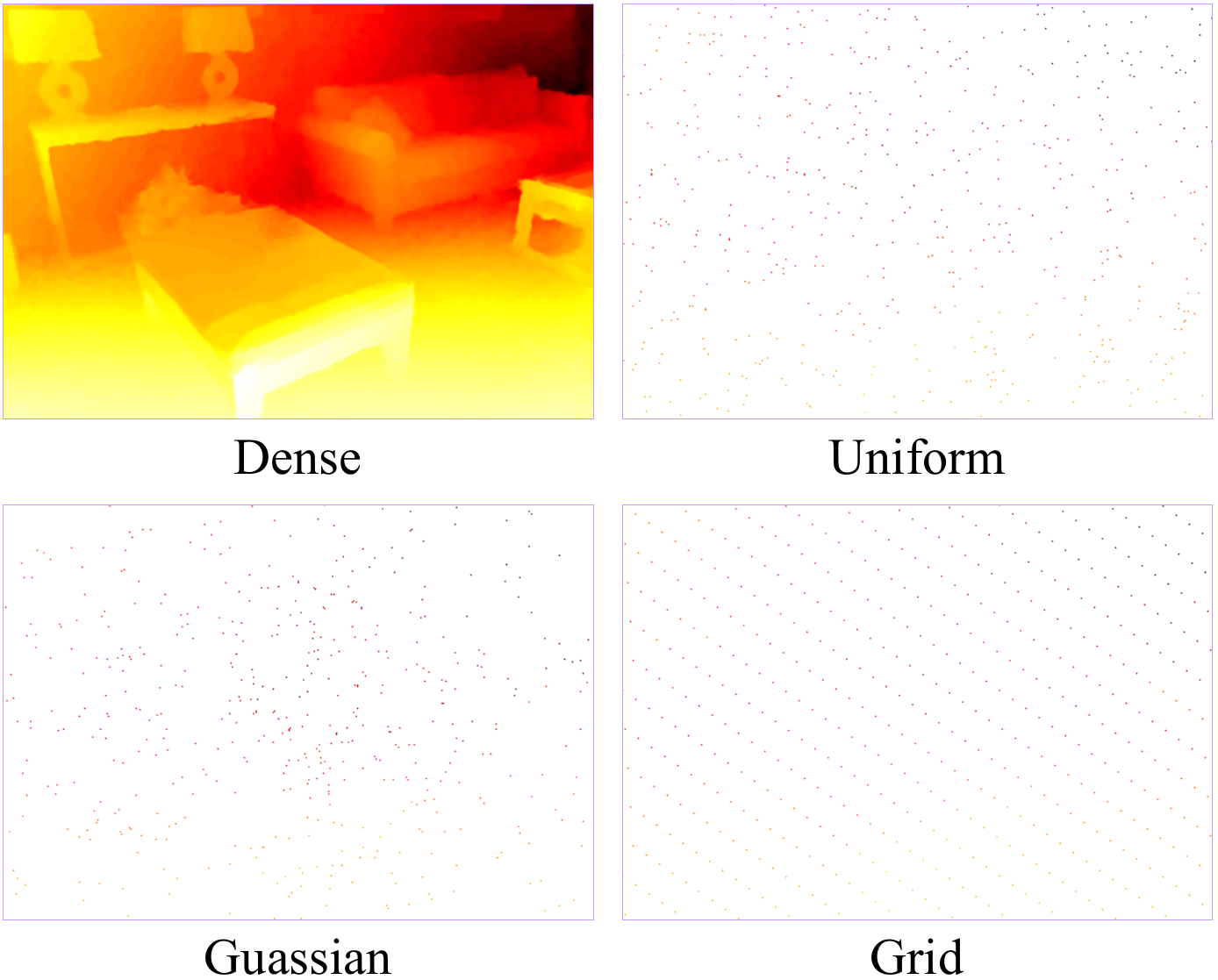}\\
  \caption{Different synthetic patterns on NYUv2 dataset.}\label{Fig_rignet_mask}
\end{figure}

\begin{table}[t]
\caption{Quantitative comparisons of different synthetic patterns on NYUv2 dataset.}
\label{density_nyu}
\centering
\large
\renewcommand\arraystretch{1.1}
\resizebox{0.4835\textwidth}{!}{
\begin{tabular}{l|l|ccccc}
\toprule
Pattern      & Method    & RMSE (m)   & REL  & ${\delta}_{1}$  & ${\delta }_{2}$ & ${\delta }_{3}$  \\ 
\midrule
\multirow{5}{*}{Uniform}   
& CSPN \cite{cspneccv}       & 0.117  & 0.016  & 99.2  & \textbf{99.9}  & \textbf{100.0} 
\\
& NLSPN \cite{park2020nonlocal}  & 0.092  & 0.012  & 99.6  & \textbf{99.9}  & \textbf{100.0}  
\\
& ACMNet \cite{zhao2021adaptive} & 0.105  & 0.015  & 99.4  & \textbf{99.9}  & \textbf{100.0}  
\\ \cmidrule{2-7}
& RigNet                         & 0.090  & 0.013  & 99.6  & \textbf{99.9}  & \textbf{100.0}  
\\ 
& \textbf{RigNet++}                       & \textbf{0.087}  & \textbf{0.010}  & \textbf{99.7}  & \textbf{99.9}  & \textbf{100.0} 
\\ \midrule
\multirow{5}{*}{Gaussian}   
& CSPN \cite{cspneccv}       & 0.121  & 0.017  & 99.1  & 99.8  & \textbf{100.0}  
\\
& NLSPN \cite{park2020nonlocal}  & 0.093  & 0.013  & 99.5  & \textbf{99.9}  & \textbf{100.0}  
\\
& ACMNet \cite{zhao2021adaptive} & 0.110  & 0.017  & 99.3  & \textbf{99.9}  & \textbf{100.0}  
\\ \cmidrule{2-7}
& RigNet                         & 0.092  & 0.012  & 99.6  & \textbf{99.9}  & \textbf{100.0}  
\\ 
& \textbf{RigNet++}                       & \textbf{0.088}  & \textbf{0.011}  & \textbf{99.7}  & \textbf{99.9}  & \textbf{100.0} 
\\ \midrule
\multirow{5}{*}{Grid}       
& CSPN \cite{cspneccv}       & 0.123  & 0.017  & 99.2  & 99.8  & \textbf{100.0}  
\\
& NLSPN \cite{park2020nonlocal}  & 0.095  & 0.013  & 99.5  & \textbf{99.9}  & \textbf{100.0}  
\\
& ACMNet \cite{zhao2021adaptive} & 0.090  & 0.012  & 99.6  & \textbf{99.9}  & \textbf{100.0}  
\\ \cmidrule{2-7}
& RigNet                         & 0.087  & 0.010  & \textbf{99.7}  & \textbf{99.9} & \textbf{100.0} 
\\ 
& \textbf{RigNet++}                       & \textbf{0.083}  & \textbf{0.009}  & \textbf{99.7}  & \textbf{99.9}  & \textbf{100.0} 
\\ 
\bottomrule
\end{tabular}
}
\end{table}

\noindent \textbf{(3) Synthetic pattern}

On NYUv2 test split, we first produce diversiform sparse depth inputs by Uniform, Gaussian, and Grid sampling manners that are shown in Fig.~\ref{Fig_rignet_mask}. Then we compare our method with three popular works with released codes and pretrained models, \emph{i.e.}, CSPN \cite{cspneccv}, NLSPN \cite{park2020nonlocal}, and ACMNet \cite{zhao2021adaptive}. Note that all models are pretrained in Uniform sampling mode and fine-tuned on those three patterns. The Uniform sampling only produces 500 valid depth points, which are consistent with the common settings of Tab.~\ref{tab_nyuv2}. As reported in Tab.~\ref{density_nyu}, \textbf{(i)} RigNet++ is superior to all other methods in the three patterns. \textbf{(ii)} In Gaussian pattern, the performance of all five methods drops a little bit, since Gaussian pattern's points around edges of sparse depth maps are fewer than those of Uniform pattern. \textbf{(iii)} In Grid pattern, which can be seen as a simple case of Uniform with regular sampling, ACMNet, RigNet, and RigNet++ perform better while CSPN and NLSPN do not. These results show RigNet can well tackle depth inputs with different sparsity patterns. 

In summary, all above-mentioned evidences verify that our method has robust generalization capabilities.

\section{Conclusion}

In this paper, we introduced a repetitive design paradigm for image guided depth completion. We identified two issues that hinder the performance of existing popular methods, namely, blurry guidance in image and unclear structure in depth. To tackle these issues, in our image guidance branch, we proposed a Dense Repetitive Hourglass Network (DRHN) to produce discriminative image features. In our depth generation branch, we designed a Repetitive Guidance (RG) module to progressively predict depth with detailed structures. Simultaneously, we incorporated the semantic prior of large vision models to enhance RG. Moreover, based on the semantic prior constraint, we developed a Region-Aware Spatial Propagation Network (RASPN) for further depth refinement. Lastly, we built a new dataset, \emph{i.e.}, TOFDC, specifically for the depth completion task. Extensive experiments have shown that our method not only delivers outstanding performance but also displays significant generalization capabilities.

\textbf{Limitation and future work}: 
(i) The semantic prior of SAM still requires additional mapping through a subnetwork. (ii) In RG, the dynamic filter relies on the simple concatenation of RGB, semantic, and depth. 
Consequently, two enhancements are imperative in our future work. 
(i) The design of feature-level prompt learning to leverage the semantic prior directly from the pre-trained model of SAM. 
(ii) The exploration of various combinations of the dynamic filter, \emph{e.g.}, initially using the RGB image to provide a coarse global context guidance, followed by the application of the semantic image to generate refined local mask guidance.


%
\section*{Conflict of interest}

The authors declare that they have no conflict of interest.


\bibliographystyle{spmpsci}      
\bibliography{egbib.bib}   

\begin{thebibliography}{100}
\providecommand{\url}[1]{{#1}}
\providecommand{\urlprefix}{URL }
\expandafter\ifx\csname urlstyle\endcsname\relax
  \providecommand{\doi}[1]{DOI~\discretionary{}{}{}#1}\else
  \providecommand{\doi}{DOI~\discretionary{}{}{}\begingroup \urlstyle{rm}\Url}\fi

\bibitem{albanis2021pano3d}
Albanis, G., Zioulis, N., Drakoulis, P., Gkitsas, V., Sterzentsenko, V., Alvarez, F., Zarpalas, D., Daras, P.: Pano3d: A holistic benchmark and a solid baseline for 360° depth estimation.
\newblock In: CVPR Workshops, pp. 3722--3732. IEEE (2021)

\bibitem{armbruster2008depth}
Armbr{\"u}ster, C., Wolter, M., Kuhlen, T., Spijkers, W., Fimm, B.: Depth perception in virtual reality: distance estimations in peri-and extrapersonal space.
\newblock Cyberpsychology \& Behavior \textbf{11}(1), 9--15 (2008)

\bibitem{cai2018cascade}
Cai, Z., Vasconcelos, N.: Cascade r-cnn: Delving into high quality object detection.
\newblock In: CVPR, pp. 6154--6162 (2018)

\bibitem{chen2023agg}
Chen, D., Huang, T., Song, Z., Deng, S., Jia, T.: Agg-net: Attention guided gated-convolutional network for depth image completion.
\newblock In: ICCV, pp. 8853--8862 (2023)

\bibitem{chen2023sam}
Chen, T., Zhu, L., Deng, C., Cao, R., Wang, Y., Zhang, S., Li, Z., Sun, L., Zang, Y., Mao, P.: Sam-adapter: Adapting segment anything in underperformed scenes.
\newblock In: ICCV, pp. 3367--3375 (2023)

\bibitem{chen2019learning}
Chen, Y., Yang, B., Liang, M., Urtasun, R.: Learning joint 2d-3d representations for depth completion.
\newblock In: ICCV, pp. 10023--10032 (2019)

\bibitem{Cheng2020CSPN}
Cheng, X., Wang, P., Guan, C., Yang, R.: Cspn++: Learning context and resource aware convolutional spatial propagation networks for depth completion.
\newblock In: AAAI, pp. 10615--10622 (2020)

\bibitem{cspneccv}
Cheng, X., Wang, P., Yang, R.: Learning depth with convolutional spatial propagation network.
\newblock In: ECCV, pp. 103--119 (2018)

\bibitem{2018Deep}
Chodosh, N., Wang, C., Lucey, S.: Deep convolutional compressed sensing for lidar depth completion.
\newblock In: ACCV, pp. 499--513 (2018)

\bibitem{cui2019real}
Cui, Z., Heng, L., Yeo, Y.C., Geiger, A., Pollefeys, M., Sattler, T.: Real-time dense mapping for self-driving vehicles using fisheye cameras.
\newblock In: ICRA, pp. 6087--6093 (2019)

\bibitem{dey2012tablet}
Dey, A., Jarvis, G., Sandor, C., Reitmayr, G.: Tablet versus phone: Depth perception in handheld augmented reality.
\newblock In: ISMAR, pp. 187--196 (2012)

\bibitem{dosovitskiy2020image}
Dosovitskiy, A., Beyer, L., Kolesnikov, A., Weissenborn, D., Zhai, X., Unterthiner, T., Dehghani, M., Minderer, M., Heigold, G., Gelly, S., et~al.: An image is worth 16x16 words: Transformers for image recognition at scale.
\newblock arXiv preprint arXiv:2010.11929  (2020)

\bibitem{dosovitskiy2017carla}
Dosovitskiy, A., Ros, G., Codevilla, F., Lopez, A., Koltun, V.: Carla: An open urban driving simulator.
\newblock In: CoRL, pp. 1--16. PMLR (2017)

\bibitem{2020Confidence}
Eldesokey, A., Felsberg, M., Khan, F.S.: Confidence propagation through cnns for guided sparse depth regression.
\newblock IEEE Transactions on Pattern Analysis and Machine Intelligence \textbf{42}(10), 2423--2436 (2020)

\bibitem{gaidon2016virtual}
Gaidon, A., Wang, Q., Cabon, Y., Vig, E.: Virtual worlds as proxy for multi-object tracking analysis.
\newblock In: CVPR, pp. 4340--4349 (2016)

\bibitem{gao2020visualechoes}
Gao, R., Chen, C., Al-Halah, Z., Schissler, C., Grauman, K.: Visualechoes: Spatial image representation learning through echolocation.
\newblock In: ECCV, pp. 658--676. Springer (2020)

\bibitem{ghiasi2019fpn}
Ghiasi, G., Lin, T.Y., Le, Q.V.: Nas-fpn: Learning scalable feature pyramid architecture for object detection.
\newblock In: CVPR, pp. 7036--7045 (2019)

\bibitem{glorot2011deep}
Glorot, X., Bordes, A., Bengio, Y.: Deep sparse rectifier neural networks.
\newblock In: AISTATS, pp. 315--323. JMLR Workshop and Conference Proceedings (2011)

\bibitem{hane20173d}
H{\"a}ne, C., Heng, L., Lee, G.H., Fraundorfer, F., Furgale, P., Sattler, T., Pollefeys, M.: 3d visual perception for self-driving cars using a multi-camera system: Calibration, mapping, localization, and obstacle detection.
\newblock Image and Vision Computing \textbf{68}, 14--27 (2017)

\bibitem{he2022masked}
He, K., Chen, X., Xie, S., Li, Y., Doll{\'a}r, P., Girshick, R.: Masked autoencoders are scalable vision learners.
\newblock In: CVPR, pp. 16000--16009 (2022)

\bibitem{He2016Deep}
He, K., Zhang, X., Ren, S., Sun, J.: Deep residual learning for image recognition.
\newblock In: CVPR, pp. 770--778 (2016)

\bibitem{howard2017mobilenets}
Howard, A.G., Zhu, M., Chen, B., Kalenichenko, D., Wang, W., Weyand, T., Andreetto, M., Adam, H.: Mobilenets: Efficient convolutional neural networks for mobile vision applications.
\newblock arXiv preprint arXiv:1704.04861  (2017)

\bibitem{hu2018squeeze}
Hu, J., Shen, L., Sun, G.: Squeeze-and-excitation networks.
\newblock In: CVPR, pp. 7132--7141 (2018)

\bibitem{hu2020PENet}
Hu, M., Wang, S., Li, B., Ning, S., Fan, L., Gong, X.: Penet: Towards precise and efficient image guided depth completion.
\newblock In: ICRA (2021)

\bibitem{huang2017densely}
Huang, G., Liu, Z., Van Der~Maaten, L., Weinberger, K.Q.: Densely connected convolutional networks.
\newblock In: CVPR, pp. 4700--4708 (2017)

\bibitem{huang2023segment}
Huang, Y., Yang, X., Liu, L., Zhou, H., Chang, A., Zhou, X., Chen, R., Yu, J., Chen, J., Chen, C., et~al.: Segment anything model for medical images?
\newblock arXiv preprint arXiv:2304.14660  (2023)

\bibitem{huang2019indoor}
Huang, Y.K., Wu, T.H., Liu, Y.C., Hsu, W.H.: Indoor depth completion with boundary consistency and self-attention.
\newblock In: ICCV Workshops, pp. 0--0 (2019)

\bibitem{huynh2021boosting}
Huynh, L., Nguyen, P., Matas, J., Rahtu, E., Heikkil{\"a}, J.: Boosting monocular depth estimation with lightweight 3d point fusion.
\newblock In: ICCV, pp. 12767--12776 (2021)

\bibitem{imran2021depth}
Imran, S., Liu, X., Morris, D.: Depth completion with twin surface extrapolation at occlusion boundaries.
\newblock In: CVPR, pp. 2583--2592 (2021)

\bibitem{ioffe2015batch}
Ioffe, S., Szegedy, C.: Batch normalization: Accelerating deep network training by reducing internal covariate shift.
\newblock In: ICML, pp. 448--456. PMLR (2015)

\bibitem{2018Sparse}
Jaritz, M., De~Charette, R., Wirbel, E., Perrotton, X., Nashashibi, F.: Sparse and dense data with cnns: Depth completion and semantic segmentation.
\newblock In: 3DV, pp. 52--60 (2018)

\bibitem{jiang2021unifuse}
Jiang, H., Sheng, Z., Zhu, S., Dong, Z., Huang, R.: Unifuse: Unidirectional fusion for 360 panorama depth estimation.
\newblock IEEE Robotics and Automation Letters \textbf{6}(2), 1519--1526 (2021)

\bibitem{kingma2014adam}
Kingma, D.P., Ba, J.: Adam: A method for stochastic optimization.
\newblock arXiv preprint arXiv:1412.6980  (2014)

\bibitem{kirillov2023segment}
Kirillov, A., Mintun, E., Ravi, N., Mao, H., Rolland, C., Gustafson, L., Xiao, T., Whitehead, S., Berg, A.C., Lo, W.Y., et~al.: Segment anything.
\newblock arXiv preprint arXiv:2304.02643  (2023)

\bibitem{ku2018defense}
Ku, J., Harakeh, A., Waslander, S.L.: In defense of classical image processing: Fast depth completion on the cpu.
\newblock In: CRV, pp. 16--22 (2018)

\bibitem{lee2021depth}
Lee, B.U., Lee, K., Kweon, I.S.: Depth completion using plane-residual representation.
\newblock In: CVPR, pp. 13916--13925 (2021)

\bibitem{levin2004colorization}
Levin, A., Lischinski, D., Weiss, Y.: Colorization using optimization.
\newblock In: SIGGRAPH, pp. 689--694. ACM (2004)

\bibitem{li2020multi}
Li, A., Yuan, Z., Ling, Y., Chi, W., Zhang, C., et~al.: A multi-scale guided cascade hourglass network for depth completion.
\newblock In: WACV, pp. 32--40 (2020)

\bibitem{li2023sam}
Li, S., Liu, M., Zhang, Y., Chen, S., Li, H., Chen, H., Dou, Z.: Sam-deblur: Let segment anything boost image deblurring.
\newblock arXiv preprint arXiv:2309.02270  (2023)

\bibitem{li2019selective}
Li, X., Wang, W., Hu, X., Yang, J.: Selective kernel networks.
\newblock In: CVPR, pp. 510--519 (2019)

\bibitem{li2022efficientformer}
Li, Y., Hu, J., Wen, Y., Evangelidis, G., Salahi, K., Wang, Y., Tulyakov, S., Ren, J.: Rethinking vision transformers for mobilenet size and speed.
\newblock In: NeurIPS, vol.~35, pp. 12934--12949 (2022)

\bibitem{li2023rethinking}
Li, Y., Hu, J., Wen, Y., Evangelidis, G., Salahi, K., Wang, Y., Tulyakov, S., Ren, J.: Rethinking vision transformers for mobilenet size and speed.
\newblock In: ICCV, pp. 16889--16900 (2023)

\bibitem{lin2023sam}
Lin, J., Liu, L., Lu, D., Jia, K.: Sam-6d: Segment anything model meets zero-shot 6d object pose estimation.
\newblock arXiv preprint arXiv:2311.15707  (2023)

\bibitem{lin2017feature}
Lin, T.Y., Doll{\'a}r, P., Girshick, R., He, K., Hariharan, B., Belongie, S.: Feature pyramid networks for object detection.
\newblock In: CVPR, pp. 2117--2125 (2017)

\bibitem{lin2022dynamic}
Lin, Y., Cheng, T., Zhong, Q., Zhou, W., Yang, H.: Dynamic spatial propagation network for depth completion.
\newblock In: AAAI, vol.~36, pp. 1638--1646 (2022)

\bibitem{lin2023dyspn_tcsvt}
Lin, Y., Yang, H., Cheng, T., Zhou, W., Yin, Z.: Dyspn: Learning dynamic affinity for image-guided depth completion.
\newblock IEEE Transactions on Circuits and Systems for Video Technology  (2023)

\bibitem{liu2021fcfr}
Liu, L., Song, X., Lyu, X., Diao, J., Wang, M., Liu, Y., Zhang, L.: Fcfr-net: Feature fusion based coarse-to-fine residual learning for depth completion.
\newblock In: AAAI, vol.~35, pp. 2136--2144 (2021)

\bibitem{liu2023mff}
Liu, L., Song, X., Sun, J., Lyu, X., Li, L., Liu, Y., Zhang, L.: Mff-net: Towards efficient monocular depth completion with multi-modal feature fusion.
\newblock IEEE Robotics and Automation Letters \textbf{8}(2), 920--927 (2023)

\bibitem{liu2017learning}
Liu, S., De~Mello, S., Gu, J., Zhong, G., Yang, M.H., Kautz, J.: Learning affinity via spatial propagation networks.
\newblock In: NeurIPS, vol.~30 (2017)

\bibitem{liu2018path}
Liu, S., Qi, L., Qin, H., Shi, J., Jia, J.: Path aggregation network for instance segmentation.
\newblock In: CVPR, pp. 8759--8768 (2018)

\bibitem{liu2022graphcspn}
Liu, X., Shao, X., Wang, B., Li, Y., Wang, S.: Graphcspn: Geometry-aware depth completion via dynamic gcns.
\newblock In: ECCV, pp. 90--107. Springer (2022)

\bibitem{liu2020cbnet}
Liu, Y., Wang, Y., Wang, S., Liang, T., Zhao, Q., Tang, Z., Ling, H.: Cbnet: A novel composite backbone network architecture for object detection.
\newblock In: AAAI, vol.~34, pp. 11653--11660 (2020)

\bibitem{2020FromLu}
Lu, K., Barnes, N., Anwar, S., Zheng, L.: From depth what can you see? depth completion via auxiliary image reconstruction.
\newblock In: CVPR, pp. 11306--11315 (2020)

\bibitem{ma2018self}
Ma, F., Cavalheiro, G.V., Karaman, S.: Self-supervised sparse-to-dense: Self-supervised depth completion from lidar and monocular camera.
\newblock In: ICRA (2019)

\bibitem{ma2018sparse}
Ma, F., Karaman, S.: Sparse-to-dense: Depth prediction from sparse depth samples and a single image.
\newblock In: ICRA, pp. 4796--4803. IEEE (2018)

\bibitem{ma2023segment}
Ma, J., Wang, B.: Segment anything in medical images.
\newblock arXiv preprint arXiv:2304.12306  (2023)

\bibitem{mazurowski2023segment}
Mazurowski, M.A., Dong, H., Gu, H., Yang, J., Konz, N., Zhang, Y.: Segment anything model for medical image analysis: an experimental study.
\newblock Medical Image Analysis \textbf{89}, 102918 (2023)

\bibitem{parida2021beyond}
Parida, K.K., Srivastava, S., Sharma, G.: Beyond image to depth: Improving depth prediction using echoes.
\newblock In: CVPR, pp. 8268--8277 (2021)

\bibitem{park2020nonlocal}
Park, J., Joo, K., Hu, Z., Liu, C.K., Kweon, I.S.: Non-local spatial propagation network for depth completion.
\newblock In: ECCV (2020)

\bibitem{qiao2021detectors}
Qiao, S., Chen, L.C., Yuille, A.: Detectors: Detecting objects with recursive feature pyramid and switchable atrous convolution.
\newblock In: CVPR, pp. 10213--10224 (2021)

\bibitem{Qiu_2019_CVPR}
Qiu, J., Cui, Z., Zhang, Y., Zhang, X., Liu, S., Zeng, B., Pollefeys, M.: Deeplidar: Deep surface normal guided depth prediction for outdoor scene from sparse lidar data and single color image.
\newblock In: CVPR, pp. 3313--3322 (2019)

\bibitem{Qu_2021_ICCV}
Qu, C., Liu, W., Taylor, C.J.: Bayesian deep basis fitting for depth completion with uncertainty.
\newblock In: ICCV, pp. 16147--16157 (2021)

\bibitem{ren2015faster}
Ren, S., He, K., Girshick, R., Sun, J.: Faster r-cnn: Towards real-time object detection with region proposal networks.
\newblock In: NeurIPS, vol.~28, pp. 91--99 (2015)

\bibitem{rey2022360monodepth}
Rey-Area, M., Yuan, M., Richardt, C.: 360monodepth: High-resolution 360deg monocular depth estimation.
\newblock In: CVPR, pp. 3762--3772 (2022)

\bibitem{rho2022guideformer}
Rho, K., Ha, J., Kim, Y.: Guideformer: Transformers for image guided depth completion.
\newblock In: CVPR, pp. 6250--6259 (2022)

\bibitem{ronneberger2015u}
Ronneberger, O., Fischer, P., Brox, T.: U-net: Convolutional networks for biomedical image segmentation.
\newblock In: MICCAI, pp. 234--241. Springer (2015)

\bibitem{shen2022panoformer}
Shen, Z., Lin, C., Liao, K., Nie, L., Zheng, Z., Zhao, Y.: Panoformer: Panorama transformer for indoor 360 depth estimation.
\newblock In: ECCV, pp. 195--211. Springer (2022)

\bibitem{silberman2012indoor}
Silberman, N., Hoiem, D., Kohli, P., Fergus, R.: Indoor segmentation and support inference from rgbd images.
\newblock In: ECCV, pp. 746--760. Springer (2012)

\bibitem{song2020channel}
Song, X., Dai, Y., Zhou, D., Liu, L., Li, W., Li, H., Yang, R.: Channel attention based iterative residual learning for depth map super-resolution.
\newblock In: CVPR, pp. 5631--5640 (2020)

\bibitem{sun2021hohonet}
Sun, C., Sun, M., Chen, H.T.: Hohonet: 360 indoor holistic understanding with latent horizontal features.
\newblock In: CVPR, pp. 2573--2582 (2021)

\bibitem{tan2019efficientnet}
Tan, M., Le, Q.: Efficientnet: Rethinking model scaling for convolutional neural networks.
\newblock In: ICML, pp. 6105--6114. PMLR (2019)

\bibitem{tan2021efficientnetv2}
Tan, M., Le, Q.: Efficientnetv2: Smaller models and faster training.
\newblock In: ICML, pp. 10096--10106. PMLR (2021)

\bibitem{tan2020efficientdet}
Tan, M., Pang, R., Le, Q.V.: Efficientdet: Scalable and efficient object detection.
\newblock In: CVPR, pp. 10781--10790 (2020)

\bibitem{tang2020learning}
Tang, J., Tian, F.P., Feng, W., Li, J., Tan, P.: Learning guided convolutional network for depth completion.
\newblock IEEE Transactions on Image Processing \textbf{30}, 1116--1129 (2020)

\bibitem{Uhrig2017THREEDV}
Uhrig, J., Schneider, N., Schneider, L., Franke, U., Brox, T., Geiger, A.: Sparsity invariant cnns.
\newblock In: 3DV, pp. 11--20 (2017)

\bibitem{vangansbeke2019}
Van~Gansbeke, W., Neven, D., De~Brabandere, B., Van~Gool, L.: Sparse and noisy lidar completion with rgb guidance and uncertainty.
\newblock In: MVA, pp. 1--6 (2019)

\bibitem{wang2021regularizing}
Wang, K., Zhang, Z., Yan, Z., Li, X., Xu, B., Li, J., Yang, J.: Regularizing nighttime weirdness: Efficient self-supervised monocular depth estimation in the dark.
\newblock In: ICCV, pp. 16055--16064 (2021)

\bibitem{wang2022cu}
Wang, Y., Dai, Y., Liu, Q., Yang, P., Sun, J., Li, B.: Cu-net: Lidar depth-only completion with coupled u-net.
\newblock IEEE Robotics and Automation Letters \textbf{7}(4), 11476--11483 (2022)

\bibitem{wang2023lrru}
Wang, Y., Li, B., Zhang, G., Liu, Q., Gao, T., Dai, Y.: Lrru: Long-short range recurrent updating networks for depth completion.
\newblock In: ICCV, pp. 9422--9432 (2023)

\bibitem{Xu2019Depth}
Xu, Y., Zhu, X., Shi, J., Zhang, G., Bao, H., Li, H.: Depth completion from sparse lidar data with depth-normal constraints.
\newblock In: ICCV, pp. 2811--2820 (2019)

\bibitem{xu2020deformable}
Xu, Z., Yin, H., Yao, J.: Deformable spatial propagation networks for depth completion.
\newblock In: ICIP, pp. 913--917. IEEE (2020)

\bibitem{xue2022go}
Xue, F., Shi, Z., Wei, F., Lou, Y., Liu, Y., You, Y.: Go wider instead of deeper.
\newblock In: AAAI, vol.~36, pp. 8779--8787 (2022)

\bibitem{yan2023distortion}
Yan, Z., Li, X., Wang, K., Chen, S., Li, J., Yang, J.: Distortion and uncertainty aware loss for panoramic depth completion.
\newblock In: ICML, pp. 39099--39109. PMLR (2023)

\bibitem{yan2022multi}
Yan, Z., Li, X., Wang, K., Zhang, Z., Li, J., Yang, J.: Multi-modal masked pre-training for monocular panoramic depth completion.
\newblock In: ECCV, pp. 378--395. Springer (2022)

\bibitem{yan2022learning}
Yan, Z., Wang, K., Li, X., Zhang, Z., Li, G., Li, J., Yang, J.: Learning complementary correlations for depth super-resolution with incomplete data in real world.
\newblock IEEE Transactions on Neural Networks and Learning Systems  (2022)

\bibitem{yan2022rignet}
Yan, Z., Wang, K., Li, X., Zhang, Z., Li, J., Yang, J.: Rignet: Repetitive image guided network for depth completion.
\newblock In: ECCV, pp. 214--230. Springer (2022)

\bibitem{yan2023desnet}
Yan, Z., Wang, K., Li, X., Zhang, Z., Li, J., Yang, J.: Desnet: Decomposed scale-consistent network for unsupervised depth completion.
\newblock In: AAAI, vol.~37, pp. 3109--3117 (2023)

\bibitem{yan2023learnable}
Yan, Z., Zheng, Y., Wang, K., Li, X., Zhang, Z., Chen, S., Li, J., Yang, J.: Learnable differencing center for nighttime depth perception.
\newblock arXiv preprint arXiv:2306.14538  (2023)

\bibitem{yang2023track}
Yang, J., Gao, M., Li, Z., Gao, S., Wang, F., Zheng, F.: Track anything: Segment anything meets videos.
\newblock arXiv preprint arXiv:2304.11968  (2023)

\bibitem{2020Denseyang}
Yang, Y., Wong, A., Soatto, S.: Dense depth posterior (ddp) from single image and sparse range.
\newblock In: CVPR, pp. 3353--3362 (2020)

\bibitem{yu20233d}
Yu, Q., Du, H., Liu, C., Yu, X.: When 3d bounding-box meets sam: Point cloud instance segmentation with weak-and-noisy supervision.
\newblock arXiv preprint arXiv:2309.00828  (2023)

\bibitem{yu2023aggregating}
Yu, Z., Sheng, Z., Zhou, Z., Luo, L., Cao, S.Y., Gu, H., Zhang, H., Shen, H.L.: Aggregating feature point cloud for depth completion.
\newblock In: ICCV, pp. 8732--8743 (2023)

\bibitem{zeiler2014visualizing}
Zeiler, M.D., Fergus, R.: Visualizing and understanding convolutional networks.
\newblock In: ECCV, pp. 818--833. Springer (2014)

\bibitem{zhang2018context}
Zhang, H., Dana, K., Shi, J., Zhang, Z., Wang, X., Tyagi, A., Agrawal, A.: Context encoding for semantic segmentation.
\newblock In: CVPR, pp. 7151--7160 (2018)

\bibitem{zhang2023customized}
Zhang, K., Liu, D.: Customized segment anything model for medical image segmentation.
\newblock arXiv preprint arXiv:2304.13785  (2023)

\bibitem{zhang2018deep}
Zhang, Y., Funkhouser, T.: Deep depth completion of a single rgb-d image.
\newblock In: CVPR, pp. 175--185 (2018)

\bibitem{zhang2023cf}
Zhang, Y., Guo, X., Poggi, M., Zhu, Z., Huang, G., Mattoccia, S.: Completionformer: Depth completion with convolutions and vision transformers.
\newblock In: CVPR, pp. 18527--18536 (2023)

\bibitem{zhang2019pattern}
Zhang, Z., Cui, Z., Xu, C., Yan, Y., Sebe, N., Yang, J.: Pattern-affinitive propagation across depth, surface normal and semantic segmentation.
\newblock In: CVPR, pp. 4106--4115 (2019)

\bibitem{zhao2017pyramid}
Zhao, H., Shi, J., Qi, X., Wang, X., Jia, J.: Pyramid scene parsing network.
\newblock In: CVPR, pp. 2881--2890 (2017)

\bibitem{zhao2021adaptive}
Zhao, S., Gong, M., Fu, H., Tao, D.: Adaptive context-aware multi-modal network for depth completion.
\newblock IEEE Transactions on Image Processing \textbf{30}, 5264--5276 (2021)

\bibitem{zhou2023bev}
Zhou, W., Yan, X., Liao, Y., Lin, Y., Huang, J., Zhao, G., Cui, S., Li, Z.: Bev@ dc: Bird's-eye view assisted training for depth completion.
\newblock In: CVPR, pp. 9233--9242 (2023)

\bibitem{zhu2019deformable}
Zhu, X., Hu, H., Lin, S., Dai, J.: Deformable convnets v2: More deformable, better results.
\newblock In: CVPR, pp. 9308--9316 (2019)

\bibitem{zioulis2019spherical}
Zioulis, N., Karakottas, A., Zarpalas, D., Alvarez, F., Daras, P.: Spherical view synthesis for self-supervised 360 depth estimation.
\newblock In: 3DV, pp. 690--699. IEEE (2019)

\end{thebibliography}

\end{document}